\documentclass[preprint,12pt]{elsarticle}

\usepackage{amssymb}
\usepackage{amsmath}
\usepackage{graphicx}

\usepackage{caption}
\usepackage{subcaption}
\usepackage{soul,color}

\begin{document}

\newcommand*{\vertbar}{\rule[-1ex]{0.5pt}{2.5ex}}
\newcommand*{\horzbar}{\rule[.5ex]{2.5ex}{0.5pt}}

\begin{frontmatter}



\title{A Distance Correlation-Based Approach to Characterize the Effectiveness of Recurrent Neural Networks for Time Series Forecasting}



\author[1]{Christopher Salazar}
\author[1,2]{Ashis G. Banerjee \corref{cor1}}

\address[1]{Department of Industrial \& Systems Engineering, University of Washington, Seattle, WA 98195, USA}
\address[2]{Department of Mechanical Engineering, University of Washington, Seattle, WA 98195, USA}

\cortext[cor1]{Corresponding author; Email address: ashisb@uw.edu}

\begin{abstract}
Time series forecasting has received a lot of attention, 
with recurrent neural networks (RNNs) being one of the widely used models 
due to their ability to handle sequential data. Previous studies on RNN time series forecasting, however, show inconsistent outcomes and offer few explanations for performance variations among the datasets.
In this paper, we provide an approach to link time series characteristics with RNN components via the versatile metric of distance correlation. 
This metric allows us to 
examine the information flow through the RNN activation layers to be able to interpret and explain their performance. 
We empirically show that the RNN activation layers learn the lag structures of time series well. However, they gradually lose this information over the span of a few consecutive layers, 
thereby worsening the forecast quality for series with large lag structures. We also show that the activation layers cannot adequately model moving average and heteroskedastic time series processes. Last, we 
generate heatmaps for visual comparisons of the activation layers for different choices of the network hyperparameters to identify which of them affect the forecast performance. Our findings can, therefore, aid practitioners in assessing the effectiveness of RNNs for given time series data without actually training and evaluating the networks. 

\end{abstract}


\end{frontmatter}


\section{Introduction}
\label{sec: Intro}

Time series forecasting is a long-standing subject that continues to challenge both academic researchers and industry professionals. Accurate forecasting of real-world operations and processes, such as weather conditions, stock market prices, 
supply chain demands and deliveries, and so on, 
are still open problems that would benefit from reliable and interpretable methods.
The challenges include 
data non-stationarity and seasonality, limited data availability with missing observations, and the inherent uncertainty of real-world events. These challenges require practitioners to have a thorough understanding of both the time series characteristics and the techniques to develop high performance models.

As the practitioners grapple with these challenges, several modeling (forecasting) techniques have been developed, which include classical statistical prediction models, such as autoregressive integrated moving average (ARIMA), statistical learning models, and deep neural networks. 
As might be expected, there has been a surge of interest in the use of deep learning models for forecasting. This is illustrated in the recent work by Zhang \textit{et al.} \cite{ZHANG2023}, where they bridge the idea of localized stochastic sensitivity to recurrent neural networks (RNNs) to induce robustness in their prediction tasks. Other works, such as the information-aware attention dynamic synergetic network \cite{HE2022143}, use multi-dimensional attention to create a novel multivariate forecasting model. The rapid emergence of diverse and complex deep learning models for forecasting has provided practitioners with numerous options. However, it has also left many of them unsure of which models to choose and how to interpret their successes and limitations.

Despite the lack of agreement in deep learning model selection, RNNs are popular candidates for time series forecasting. This is due to their ability to handle sequential data, which is a fundamental characteristic of time series. Yet, through a series of several experiments, RNNs and their adaptations \cite{lai2018modeling, huang2019dsanet, zhou2021informer, lim2021time} have shown a wide range of forecasting results, making it difficult to evaluate their effectiveness. Experimental reviews and surveys have highlighted both the scenarios, one in which RNNs have achieved highly accurate results \cite{lara2021experimental, sezer2020financial, gasparin2022deep, yan2018financial, wegayehu2022short}, and another in which they have been surpassed by traditional statistical regression models such as ARIMA \cite{bousqaoui2021comparative} and gradient boosting trees \cite{elsayed2021we}. Despite the breadth of these studies, their conclusions offer limited insights on the variations in RNN performance, often deferring to the explainability of the chosen hyperparameters or domain-specific hypotheses. These outlooks can be attributed to the restricted perspective offered by generalization error, which is the primary way of evaluating these models. While the generalization error is important to track, it does not delve into the internal mechanism of how time series inputs evolve through the RNN layers, limiting our understanding of this subject. As it stands, an evaluation tool that allows us to comprehend the inner workings of RNNs on a layer-by-layer basis is needed. 

While there are several evaluation metrics to measure mathematical and statistical relationships, many lack the flexibility that is needed for studying the components of RNNs. A recently proposed statistical dependency metric, called distance correlation \cite{szekely2013distance}, 
has a few key advantages 
in this regard. First, it has the ability to capture the non-linear relationships 
present in typical RNN operations. Additionally, it can compare random variables of different dimensions, which are likely to occur in RNN architectures. It can also be used as a proximal metric to measure the information stored in the RNN activation layers through its training cycle \cite{zhen2022versatile, vepakomma2020nopeek}.
These factors make distance correlation well suited for a detailed characterization of RNN models based on their forecast performance.

\begin{figure}[!ht]
    \centering
		\includegraphics[width = \textwidth]{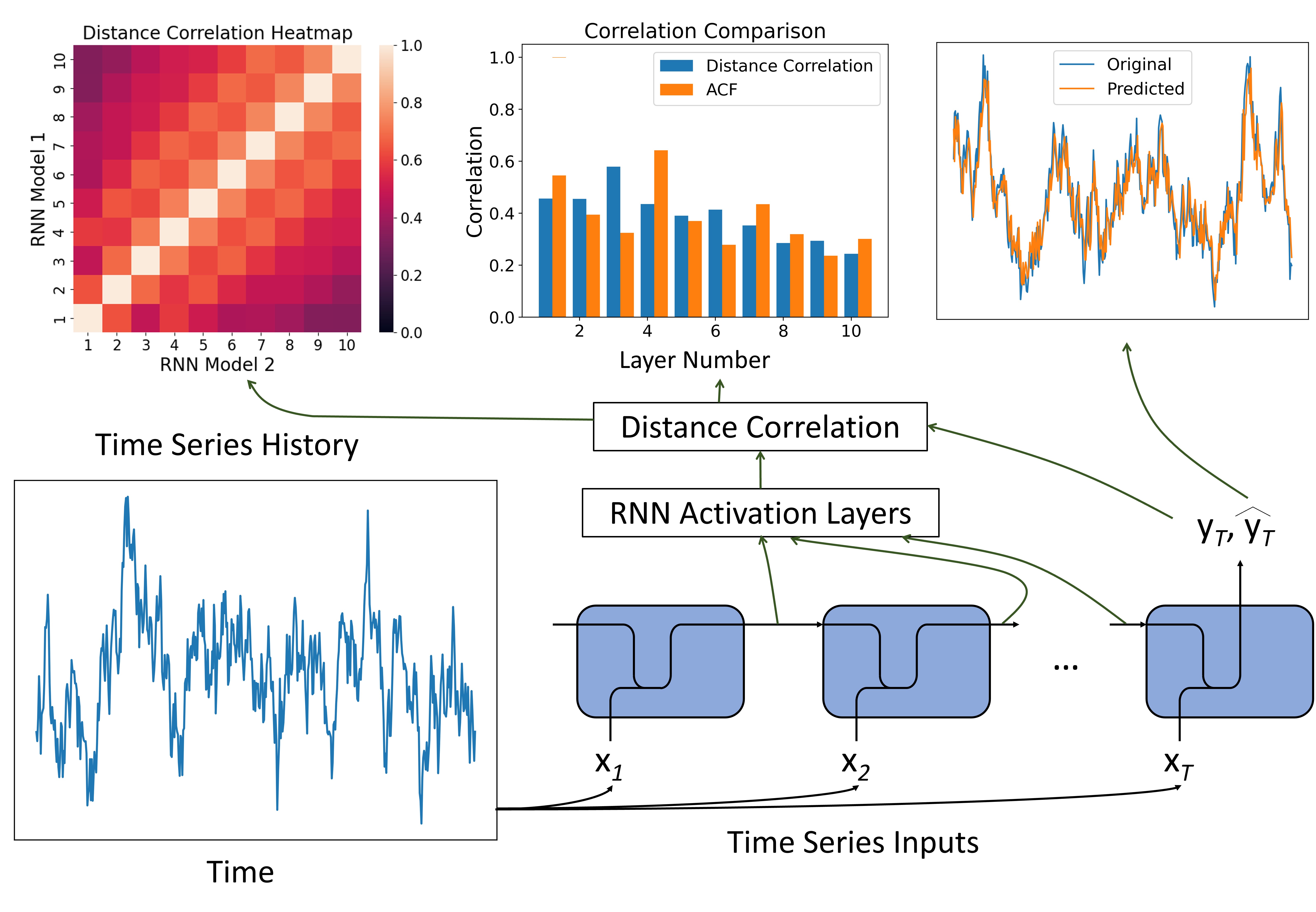}
	\caption{Overview of the use of distance correlation to examine time series forecasting using a recurrent neural network (RNN). We begin with a time series history that comprises the inputs $x_{t}$ for the RNN and the predicted outputs $\hat{y}_{T}$. The outputs for each activation layer and ground truth values, $y_{T}$, are extracted and processed with distance correlation. This is then used to generate correlation plots for analyzing activation layers behaviors and visualizing heatmaps for comparisons of different RNN models.}
	\label{fig:overall_meth}
\end{figure}

Given this context, we make the following contributions to the deep learning-based time series forecasting literature.
First, we outline a distance correlation-based approach for evaluating and understanding time series modeling 
at the RNN activation layer level. 
Consequently, we demonstrate that the activation layers 
identify time series lag structures well. However, they quickly lose this important information over a sequence of a few layers, posing difficulties in modeling time series with large lag structures.  
Further, our experiments show that the activation layers cannot adequately model moving average and heteroskedastic processes.
Last, we employ distance correlation to visualize heatmaps that compare RNNs with varying hyperparameters. These heatmaps show that certain hyperparameters, such as the number of hidden units and activation function, do not influence the activation layer outputs as much as the RNN input size. A visual overview of our work is shown in Figure \ref{fig:overall_meth}. 

\section{Related Works}
\label{sec: rela_work}

Neural networks, despite their tremendous success, are not fully understood from a theoretical perspective. Some strides have been made, such as the work by Telgarsky \cite{telgarsky2016benefits}, where the greater approximation power of deeper networks is proved for a class of nodes that includes ReLU and piecewise polynominal functions. Rolnick's work \cite{rolnick2018towards} extended this rigorous understanding of neural networks by establishing the essential properties of expressivity, learnability, and robustness. Bahri \textit{et al.} \cite{bahri2020statistical} recently reviewed a breadth of research 
on the connections between deep neural networks and several principles of statistical mechanics to obtain a better conceptual understanding of deep learning capabilities. To estimate the overall complexity and learning ability of deep neural networks without actual training and testing, 
Badias and Banerjee proposed a new layer algebra-based framework to measure the intrinsic capacity and compression of networks, respectively \cite{badias2023neural}. These measurements enabled us to analyze the performance (accuracy) of state-of-the art networks on benchmark computer vision datasets. 

Specifically for RNNs, one of the more well-known drawbacks is their inability to retain information from long term sequences, as found by Bengio \textit{et al.} while using gradient-based algorithms for training RNNs \cite{bengio1994learning}. 
This finding led to the use of alternate RNN architectures, such as the Long Short-Term Memory (LSTM) network \cite{hochreiter1997long}, despite their increased computational complexity. 
More recently, Chen \textit{et al.} \cite{chen2019generalization} established generalization bounds for RNNs and their variants based on the spectral norms of the weight matrices and the network size.  
A different approach 
focused on studying the implicit regularization effects of RNN by injecting noise into the hidden states, which resulted in promoting flatter minima and overall global stability of RNNs \cite{lim2021noisy}. While all these approaches are theoretically rigorous, 
they are typically formulated under rigid constraints, many of which are not necessarily satisfied in real-world applications. 

Therefore, several researchers have turned to interpretation based methods to develop a principled yet practically useful understanding of RNNs. In this context, since RNNs are common for natural language processing (NLP) tasks, many studies have focused on tracking the hidden memories of RNNs and their expected responses to the input text \cite{ming2017understanding, strobelt2017lstmvis}. For time series, Shen \textit{et al.} \cite{shen2020visual} developed a visual analytics system to interpret RNNs for high dimensional (multivariate) forecasting tasks. While this system is very useful, we aim to approach RNN interpretability from a different perspective of understanding performance (forecast) generalizability by tracking how the hidden memories respond to specific 
time series characteristics.
For this purpose, we look at alternate information-theoretic approaches.

One of the most promising works on information-theoretic analysis 
was conducted by Schwartz and Tishby \cite{shwartz2017opening}. They used mutual information (MI) 
to represent a deep neural network as a series of encoder-decoder networks, enabling analysis at the activation layer level. Such analysis goes beyond the typical generalization errors used for model assessment, and allows us to closely observe the information flow through deep learning networks such as RNNs.
Although this approach is novel, 
MI estimation is challenging for real-world data. 
To address this challenge, researchers have investigated different approaches, such as 
a $k$-nearest neighbor method \cite{kraskov2004estimating} and a semi-parametric local Gaussian method \cite{gao2015efficient}. While these attempts 
seem promising, there are questions regarding their practical usefulness. 
For example, McAllester and Stratos \cite{mcallester2020formal} argue that without any known probability characteristics of the data, any distribution-free, high-confidence lower bound on mutual information would require an exponential amount of samples. 
It is quite rare to find real world datasets with known (or well-fitting) distributions, 
and this is generally 
true for time series applications \cite{brockwell_1996}. Nevertheless, 
the underlying framework of analyzing RNNs from a layer-by-layer perspective is a useful concept 
that we can build upon using an alternative metric, as discussed next.

As we search for an approach 
that affords the flexibility of analyzing the components of RNNs, Szekely \textit{et al.'s} \cite{szekely2007measuring} distance correlation comes to the forefront as a dependency measure. It is closely related to Pearson's correlation with the advantage of measuring non-linear relationships among the random variables. Naturally, some researchers directly apply it toward time series forecasting. Zhou's \cite{zhou2012measuring} work re-purposes distance correlation by proposing the auto-distance correlation function (ADCF), which is an extension of the auto-correlation function (ACF) \cite{brockwell_1996}. Yousuf and Feng \cite{yousuf2022targeting} use distance correlation as a screening method for deciding which lagged variables should be retained in a model. 

Beyond time series applications, distance correlation has other practical uses in the deep learning space. Zhen \textit{et al.} \cite{zhen2022versatile} outline a process that uses distance correlation and partial distance correlation to compare deep learning models from a network layer level. They use it as a measure of information lost or gained in the network layers, where high distance correlation values indicate more information is stored in the activation layers. In a separate study, NoPeek \cite{vepakomma2020nopeek} uses a distance correlation method that decorrelates the raw input information within the activation layers during training, which is essential for preventing the leakage of sensitive data. These studies show the viability of distance correlation as an analysis tool in time series applications and deep learning, separately. Our work aims to connect the two topics to further our 
understanding of time series forecasting tasks.

\section{Methodology}

In this section, we first characterize the time series forecasting problem before describing the components of an RNN and providing an overview of distance correlation. We then propose a method that links these core topics to further our understanding of time series forecasting using RNNs.

 \subsection{Time Series Forecasting}\label{sec:TS_Forecast}

As we define the forecasting problem, we adopt the notation used by Lara-Benitez et al. \cite{lara2021experimental} in their experimental review of deep learning models with forecasting tasks. We let $\textbf{Z} = z_{1}, z_{2}, \dots, z_{L}$ be the historical time series data of length $L$ and $H$ be the prediction horizon. The goal of a time series forecasting problem is to use $\textbf{Z}$ to accurately predict the next horizon values $z_{L+1}, z_{L+2}, \dots, z_{L+H}$. The time series data can either be univariate or multivariate; we focus this paper on univariate analysis where single and sequential observations are recorded over time.

One of the challenges of time series research is the lack of benchmark datasets for forecasting. Many time series experimental reviews \cite{lara2021experimental, yan2018financial, wegayehu2022short} choose datasets from several domains with various characteristics. This 
makes it difficult to generalize the behavior of the forecasting models since it is not clear whether each dataset represents a class of time series processes or is uniquely domain specific. It also complicates the interpretability of the experiments and their corresponding conclusions. 
To circumvent these issues, we experiment primarily with synthetic data from known time series processes (see Section \ref{sec: Gen_TS}).
We, however, still use real world time series data to provide a practical context for our proposed forecasting model evaluation tool.

\begin{figure}
    \centering
		\includegraphics[width = 0.7\textwidth]{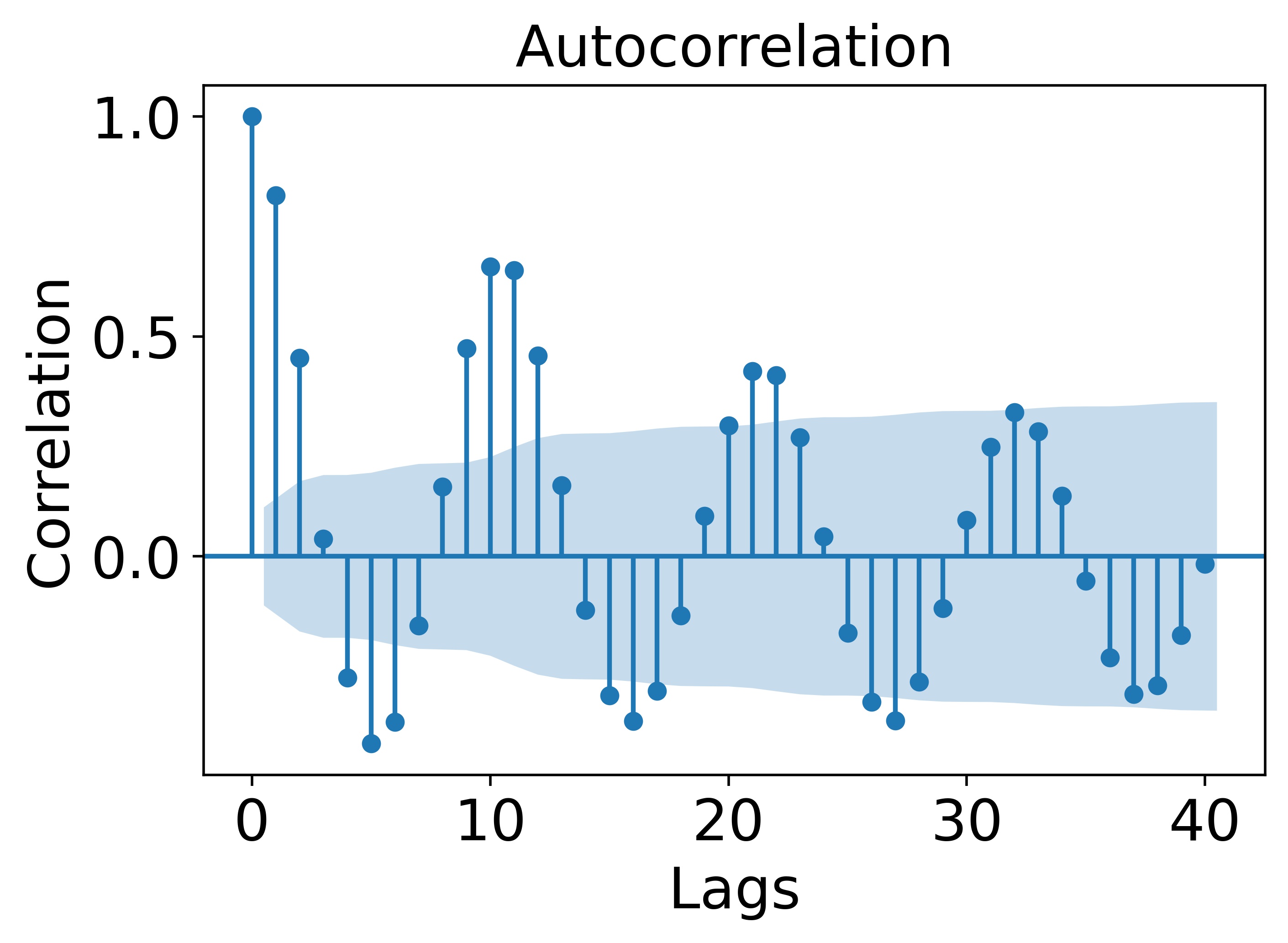}
	\caption{An example of an auto-correlation plot for the sun spot time series data \cite{center_2009} with blue shaded significance level band. Lags that fall outside of significance level band are considered important to include in a forecasting model. It also provides a way to describe the time series data, as the cyclical nature of the lags from this plot indicates a seasonal characteristic to sun spot observations.}
	\label{fig:acf_ex}
\end{figure}

Given our data, it is important for us to establish a baseline of the time series structure. One of the primary methods for analyzing time series forecasting problems is by using the auto-correlation function (ACF) \cite{brockwell_1996}. This function looks at the linear correlations between $z_{l}$ and its previous values $z_{l-h}$ where $l = 1,2 \ldots, L$ at lag $h$. Formally, the auto-correlation function, $\rho_z(h)$, is defined as 
\begin{align}\label{eq: pcor}
    \rho_{z}(h) = Cor(z_{l}, z_{l-h})
\end{align}
\noindent where $Cor$ represents the Pearson's correlation function. Plotting the ACF is a common method for understanding a time series lag structure. An example of this auto-correlation plot is shown in Figure \ref{fig:acf_ex} for the sun spot time series data \cite{center_2009}. 

A final key aspect to consider is the size of the prediction horizon $H$. In this study, we stick with single-step ahead forecasting ($H = 1$) to create a simple experimental environment. This choice allows us to minimize the model complexity required for multi-step forecasting and reduces the error accumulation with having a larger $H$. 

\subsection{RNN Architecture} 
\label{sec: RNN_Arch}

\begin{figure}
	\centering
		\includegraphics[width=\textwidth]{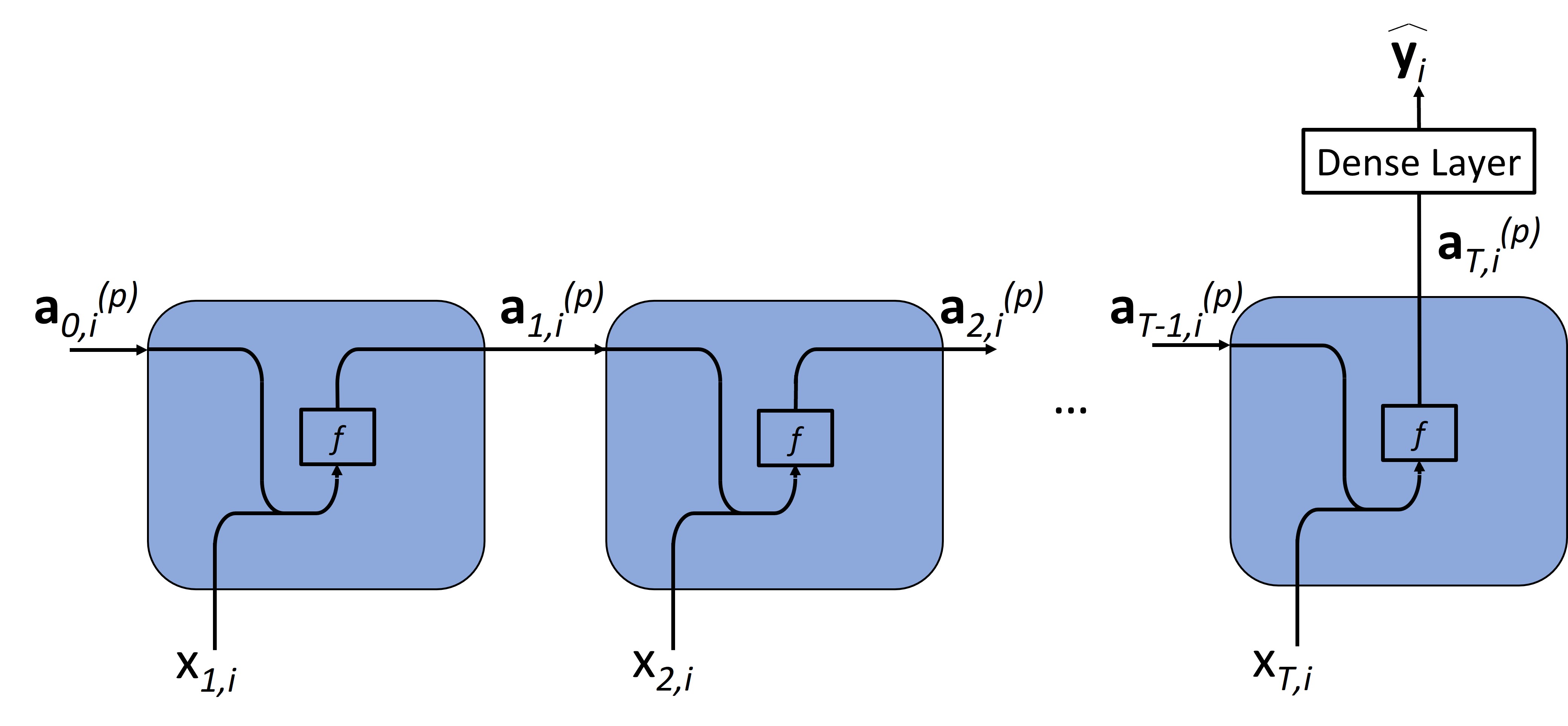}
	\caption{Time series input-output structure with an unfolded RNN for the $i$th sample and training epoch $p$. The $i$th input of the RNN  $\textbf{x}_{i} = [x_{1,i}, x_{2,i}, ... , x_{T,i}]$ produces activation layer outputs $\textbf{a}_{t,i}^{(p)}$ for $t = 0, \dots, T$ using a predetermined activation function $f$. The final activation layer output $\textbf{a}_{T,i}^{(p)}$ is fed into a dense layer to produce a single-step ahead forecast value $\hat{\textbf{y}}_{i} = \{\hat{x}_{T+H,i}\}$ for $H = 1$.}
	\label{fig:unfolded_rnn}
\end{figure}

Considering that we are 
investigating recurrent neural networks (RNNs), we must define all its components. RNN was originally proposed by Elman \cite{elman1990finding}, where they altered the classical feedforward network layer structure into a recurrent layer. This change in architecture allowed the outputs of a previous layer to become the inputs of the current layer, thereby creating a compact model representation for sequential input data. Figure \ref{fig:unfolded_rnn} shows an unfolded RNN displaying the inputs, an activation function operation (i.e., tanh), and outputs (hidden state). While the Elman RNN is relatively simple compared to other model architectures, focusing on this network provides a baseline of how recurrent based networks learn time series structures.

\begin{figure}[!ht]
	\centering
		\includegraphics[width=\textwidth]{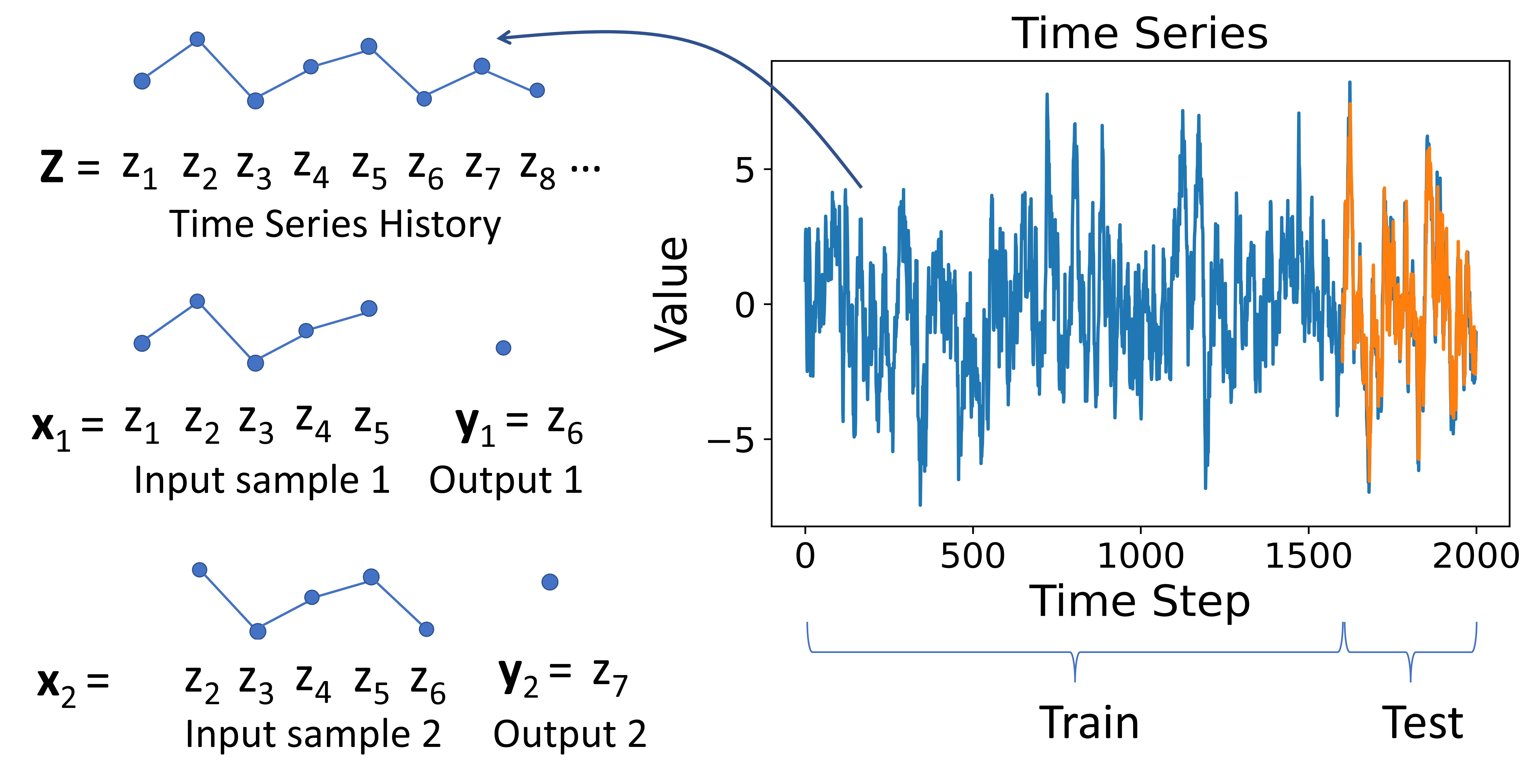}
	\caption{Time series sampling strategy where a full univariate time series plot is displayed. This full time series is partitioned into an 80:20 training:test split. Each of these training and testing splits are further divided into input-output samples, where each sample is generated via a sliding window. In this example, our sliding window is of size $T = 5$ and prediction horizon $H = 1$.} 
	\label{fig:ts_samp}
\end{figure}

A necessary step for modeling time series with an RNN is preparing the full univariate time series history into samples that purposefully generate the input-output sequences. We adopt the moving window procedure described by \cite{lara2021experimental} to create the input and output samples. Given a full time history $\textbf{Z}$, 
a window of fixed size $T$ slides over the 
data to create samples of input $\textbf{x}_{i}$ in $\mathbb{R}^{T}$ and a corresponding output $\textbf{y}_{i}$ in $\mathbb{R}^{H}$, where $i = 1, 2, \ldots, n$. A visual example of generating these input-output samples 
is shown in Figure \ref{fig:ts_samp} for $T = 5$ and $H = 1$.

The input-output sample vectors, $(\textbf{x}_{i},\textbf{y}_{i})$, are currently globally represented in terms of their $z_{l}$ values. However, it is beneficial to convert our vectors into a sample representation that serve as local inputs for our RNN model. To do this, we 
denote our samples as $\textbf{x}_{i} = [x_{1,i}, x_{2,i}, ... , x_{T,i}] $, which represents the \textit{i}th sample of a sequence of $T$ time series values. For each component of $\textbf{x}_{i}$, we recover its relation to the time series history using $x_{t,i} = z_{t+i-1}$, where $t = 0, 1, \ldots, T$. Similarly, the output of the RNN is a vector $\hat{\textbf{y}}_{i} = [\hat{x}_{T+1,i}]$, which represents the \textit{i}th forecasting output for a single-step ahead value beyond $T$, or $H = 1$. We measure the output error with reference to the ground truth single-step ahead vector $\textbf{y}_{i} = [x_{T+1,i}]$. 

Lastly, we define the activation layer outputs (or hidden states) in the context of our time series data and the RNN architecture. Considering an input-output sample $(\textbf{x}_{i}, \textbf{y}_{i})$, the hidden state $\textbf{a}_{t,i}^{(p)}$ represents the $i$th sample output of an activation layer at time step $t$ and epoch $p$, where $p = 1, 2, \dots, P$. The activation layer output $\textbf{a}_{t,i}^{(p)}$ is a vector in $\mathbb{R}^{b}$, where $b$ specifies the width of the hidden state or the number of hidden units. Each activation layer output is produced from tensor operations with the sample inputs that pass through some activation function $f$. The time step $t$ for the activation layer output $\textbf{a}_{t,i}^{(p)}$ doubles as an indicator for the layer number. For example, $t = 2$ corresponds to the input $x_{2,i}$ and $\textbf{a}_{2,i}^{(p)}$, the second activation layer for training epoch $p$. In order to output a single prediction value, we must collapse the dimension of the final activation layer output, $\textbf{a}_{T,i}^{(p)}$, with a dense layer. A visualization of this unrolled RNN setup with the time series forecasting problem is shown in Figure \ref{fig:unfolded_rnn}.

\subsection{Distance Correlation}
\label{sec: EnergyStat}

An approach for examining the relationships among random variables is through the use of energy statistics \cite{szekely2013energy}. Energy statistics comprise a set of functions derived from the distances of statistical observations, inspired by the gravitational potential energy between two bodies. We primarily use the  distance correlation metric from energy statistics \cite{szekely2007measuring} to measure the statistical dependence between random variables. Distance correlation is tailored to measure the non-linear relationships between two random variables, not necessarily of equal sizes. We build the empirical definition of distance correlation by relating it some of our previously defined time series components. We start with our input-output samples as $\mathbf{X} \in \mathbb{R}^{T \times n}$  and $\mathbf{Y} \in \mathbb{R}^{H \times n}$: 
\[
\mathbf{X} = 
\left[
  \begin{array}{cccc}
    \vertbar & \vertbar &        & \vertbar \\
    \mathbf{x}_{1}    & \mathbf{x}_{2}    & \ldots & \mathbf{x}_{n}    \\
    \vertbar & \vertbar &        & \vertbar 
  \end{array}
\right], 
\mathbf{Y} = 
\left[
  \begin{array}{cccc}
    \vertbar & \vertbar &        & \vertbar \\
    \mathbf{y}_{1}    & \mathbf{y}_{2}    & \ldots & \mathbf{y}_{n}    \\
    \vertbar & \vertbar &        & \vertbar 
  \end{array}
\right]
\] 
\noindent where the columns $\mathbf{x}_{i} \in \mathbb{R}^{T}, \mathbf{y}_{i} \in \mathbb{R}^{H}$ are the input-output sample vectors defined in section \ref{sec: RNN_Arch}. From the observed samples $\mathbf{X}, \mathbf{Y}$, define 
\[
b_{ij} = \|\mathbf{x}_{i} - \mathbf{x}_{j}\| \quad \overline{b}_{i.} = \left(\sum_{j = 1}^{n}b_{ij}\right)/n \quad \overline{b}_{.j} = \left(\sum_{i = 1}^{n}b_{ij}\right)/n
\]
\[
\overline{b}_{..} = \left(\sum_{i,j = 1}^{n}b_{ij}\right)/n^{2} \quad B_{ij} = b_{ij} - \overline{b}_{i.} - \overline{b}_{.j} + \overline{b}_{..}
\]
\noindent These are similarly defined for $\overline{c}_{i.}$, $\overline{c}_{.j}$, $\overline{c}_{..}$ where $c_{ij} = \|\mathbf{y}_{i} - \mathbf{y}_{j}\|$ and $C_{ij} = c_{ij} - \overline{c}_{i.} - \overline{c}_{.j} + \overline{c}_{..}$. Given this set of double centered matrices, the sample distance covariance function \cite{szekely2007measuring} is defined as
\begin{align}
    \hat{V}^{2}(\mathbf{X}, \mathbf{Y}) = \frac{1}{n^{2}} \sum_{i,j = 1}^{n} B_{ij}C_{ij} \quad.
\end{align}

\noindent Finally, the empirical distance correlation is defined as 
\begin{align}\label{eq:emp_dist}
    \hat{R}^{2}(\mathbf{X},\mathbf{Y}) =  
\begin{cases}
    \frac{\hat{V}^{2}(\mathbf{X},\mathbf{Y})}{\sqrt{\hat{V}^{2}(\mathbf{X})\hat{V}^{2}(\mathbf{Y})}},& \hat{V}^{2}(\mathbf{X})\hat{V}^{2}(\mathbf{Y}) > 0\\
    0,              & \hat{V}^{2}(\mathbf{X})\hat{V}^{2}(\mathbf{Y}) = 0
\end{cases}
\end{align}

\noindent The empirical distance correlation function has the following properties \cite{edelmann2019updated}: 
\begin{itemize}
  \item $\hat{R}(\mathbf{X},\mathbf{Y})$ converges almost assuredly to the theoretical counterpart \cite{szekely2007measuring} provided $n \rightarrow \infty$ and $E[\|\mathbf{X}\|] < \infty$ and $E[\|\mathbf{Y}\|] < \infty$. 
  \item $\hat{R}(\mathbf{X},\mathbf{Y}) = 0$ denotes the independence of $\mathbf{X}$ and $\mathbf{Y}$ 
  \item 
  $0 \leq \hat{R} \leq 1$
\end{itemize}
\noindent Additionally, another key property shown by \cite{szekely2013distance}, for a fixed $n$
\begin{align}\label{eq: dcor_inf}
    \lim_{T,H \to \infty} \hat{R}(\mathbf{X}, \mathbf{Y}) = 1    
\end{align}
\noindent provided that $\mathbf{X}$ and $\mathbf{Y}$ are independently and identically distributed (i.i.d.) and their corresponding second moments exist. 

\subsection{Analysis of RNN Activation Layers}\label{sec: meas_dist_RNN}

We now provide the framework for using distance correlation to study how the RNN activation layer outputs learn time series structures that eventually translate to 
forecasting outputs. We start with a set of samples $(\textbf{x}_{1}, \textbf{y}_{1})$, $(\textbf{x}_{2}, \textbf{y}_{2})$, ... , $(\textbf{x}_{n}, \textbf{y}_{n})$ where each $\textbf{x}_{i}$ is an RNN input to produce a forecasting value $\hat{\textbf{y}}_{i}$. This is compared with the ground truth output $\textbf{y}_{i}$ using a loss function. For each sample fed into the RNN during training, a unique activation layer output $\textbf{a}_{t,i}^{(p)}$ is produced depending on the sample $i$, time step $t$, and epoch $p$. The accumulation of all the samples for a particular activation layer output at time step $t$ and epoch $p$ is represented as a matrix
\[
\textbf{A}_{t}^{(p)} = 
\left[
  \begin{array}{cccc}
    \vertbar & \vertbar &        & \vertbar \\
    \textbf{a}_{t,1}^{(p)}    & \textbf{a}_{t,2}^{(p)}    & \ldots & \textbf{a}_{t,n}^{(p)}    \\
    \vertbar & \vertbar &        & \vertbar 
  \end{array}
\right]
\]
\noindent With our representation of sample ground truth outputs $\mathbf{Y}$, we estimate the distance correlation between the activation layer output at time $t$ and epoch $p$, and the ground truth outputs. This is denoted by $\hat{R}(\textbf{A}_{t}^{(p)}, \mathbf{Y})$, and is calculated using (\ref{eq:emp_dist}). We also estimate the dependency between two activation layers, $\textbf{A}_{v}^{(p)}, \textbf{A}_{m}^{(p)}$, at different time steps using $\hat{R}(\textbf{A}_{v}^{(p)}, \textbf{A}_{m}^{(p)})$, which is the distance correlation between the activation layer outputs at time steps $v$ and $m$ at epoch $p$. We note that for both the scenarios, the dimensions between the two random variables need not be the same. This framework guides the experiments 
in section \ref{sec: Experiments}. 

\section{Experiments} \label{sec: Experiments}

We 
now conduct a series of experiments to showcase the usefulness of our method for assessing time series forecasting using RNN models. Using such method, we make note of certain limitations of RNN forecasting and compare the models (network architectures) using heatmap visualizations.  

\subsection{Time Series Data}\label{sec: Gen_TS}

Before we begin our experiments, we must select the type of time series data we would like to analyze. As discussed in section \ref{sec:TS_Forecast}, we generate synthetic time series data for each time step $z_{l}$, according to a few well established processes. Perhaps the simplest case is the auto-regressive process, signified by AR($p$), where the value of the current time step $l$ value is a linear combination of the previous $p$ time steps. Formally, the auto-regressive process is defined as \cite{brockwell_1996}
\begin{align}\label{eq: AR}
    z_{l} = \sum_{i = 1}^{p} c_{i}z_{l-i} + \epsilon_{l} \quad.
\end{align}
\noindent Here, $c_{i}$ are the coefficients of the lagged variables and $\epsilon_{l}$ is white noise which follows a normal distribution $N(\mu, \sigma^{2})$ at time step $l$. We also consider the moving average model, MA($q$), which is similarly defined as \cite{brockwell_1996}
\begin{align}\label{eq: MA}
    z_{l} =  \delta + \sum_{i = 1}^{q} \theta_{i}\epsilon_{l-i}
\end{align}
\noindent where $\theta_{i}$ are the coefficients of the lagged white noised $\epsilon_{l-i}$ and $\delta$ is a predetermined average value. This moving average process generates the value of the current time step $l$ based on an average value $\delta$ and the previous $q$ lagged white noise. When we combine the processes from (\ref{eq: AR}) and (\ref{eq: MA}), we arrive at the auto-regressive moving average, or the ARMA($p,q$) model. 

Beyond these simple models, we consider another time series model called the generalized autoregressive conditional heteroskedasticity (GARCH) model. Full definition of the GARCH($p,q$) process is presented in \cite{brockwell_1996}, where this process allows the time series to exhibit variance spikes during any given time step. It is generally defined as: 
\begin{align}\label{eq: GARCH_Zt}
    z_{l} = \sqrt{h_{l}}\epsilon_{l}
\end{align}
\noindent where $h_{l}$ is a positive function that represents the variance. The variance $h_{l}$ is defined by: 
\begin{align}\label{eq: GARCH_ht}
    h_{l} = \alpha_{0} + \sum^{p}_{i = 1}\alpha_{i}z^{2}_{l-i} + \sum^{q}_{j = 1}\beta_{j}h^{2}_{l-j} \quad.
\end{align}
\noindent In this GARCH($p,q$) definition, 
$\alpha_{i}$ are the coefficients of the $p$ lagged variables, and $\beta_{j}$ are the coefficients of the $q$ lagged variances. This process induces volatility in the time series data that is often present in real world datasets. One simple but effective way to characterize these times series processes is by observing the magnitudes of their lag structures. For large values of $p,q$, we consider the lag structure to be high (and correspondingly low for small values of $p,q$). 

In addition to the synthetic time series processes, we also experiment with the following three real world datasets. 
(1) ETTh1, OT \cite{zhou2021informer} includes the hourly oil temperature (OT) collected by electricity transformers from July 2016 to July 2018 from two different counties in China. (2) Solar-energy \cite{lai2018modeling} includes the solar power production records in the year of 2006, which is sampled every 10 minutes from 137 photovoltaic plants in Alabama, USA. (3) Daily NASDAQ composite index values between 2014 to 2023 are gathered from Yahoo! Finance \cite{Finance_2024}. Note that ETTh1, OT, and Solar-energy have served as benchmarks for time series forecasting tasks \cite{lai2018modeling, liu2022non, zeng2023transformers}.

\subsection{RNN behavior under various time series processes}\label{sec: RNN_limits}

The focus of this section is to use distance correlation to determine the characteristics of RNNs for various time series processes. To investigate this, we carry out a series of experiments that track the outputs of each activation layer and compare it to the ground truth output via distance correlation upon the completion of model training. In effect, we are calculating $\hat{R}(\textbf{A}_{t}^{(P)}, \textbf{Y})$ for $t = 1,2, ..., T$ at the final epoch, $P$. $\hat{R}(\textbf{A}_{t}^{(P)}, \textbf{Y})$ values near one show that the activation layer at $t$ has a strong dependency to $\textbf{Y}$, indicating it is a highly important layer in the training process. Conversely, values close to zero imply that the activation layer at $t$ does not contribute to learning the output $\textbf{Y}$. 

One particular component of interest is layer number $T$, $\textbf{A}_{T}^{(P)}$, which is the final activation layer before the RNN generates an output. Since all the previous inputs flow to this activation layer, tracking $\hat{R}(\textbf{A}_{T}^{(P)}, \textbf{Y})$ serves as a measure of what information has been retained or lost during training.

Since we are tracking the dependency structure of each layer $\textbf{A}_{t}^{(P)}$ and ground truth $\textbf{Y}$, we can make a parallel comparison to the auto-correlation function (ACF) defined by 
(\ref{eq: pcor}). If we are interested in understanding the linear structure between a single-step ahead time series value and a corresponding lag $h$, we would calculate  $Cor(z_{l+1}, z_{l+1-h})$. The analogous comparison to this would be calculating the distance correlation, $\hat{R}(\textbf{A}_{T+1-h}^{(P)}, \textbf{Y})$. For example, given a window size of $T = 20$ and a lag of $h = 5$, activation layer number 16, $\textbf{A}_{16}^{(P)}$, can be directly compared to the ACF value at the 5th lag. Note that this inverse relationship between the layer number and lag (e.g., layer 20 and lag 1, or, layer 1 and lag 20) is used for comparisons during our experiments. We make this comparison to establish a baseline analysis of the underlying time series structure.

\subsubsection{Implementation Details}
                     
Before presenting the results, we choose some parameters specific to our experiments. We begin by generating a time series history $Z$ of length $L = 4,000$ from the processes and datasets in Section \ref{sec: Gen_TS}.
We change $L$ to 2,000 and 1,500 for the Solar-energy and NASDAQ datasets, respectively, to accommodate their smaller size. As is customary in training deep learning models, we standardize each time series using the z-score \cite{lara2021experimental}. This is done to aid the stability of training an RNN and analyze all the time series processes or datasets using the same scale. We then split our time series history such that the first 80$\%$ points form the training set and the last 20$\%$ points are used for evaluation (test set), as shown in Figure \ref{fig:ts_samp}. We only consider single step-ahead forecasts, i.e., $H = 1$. 

For the Elman RNN setup, we fix the input window size $T = 20$, the number of hidden units as $b = 64$, use the ReLU activation function, batch size of 64, learning rate of 0.0001 and train the model for 35 epochs. We also initialize the networks weights via a Kaiming He initialization procedure \cite{he2015delving}, as it has been shown to be a stable and efficient initialization scheme \cite{rolnick2018towards}. We choose these parameters with the aim of creating a stable environment where the RNN has a high likelihood of converging. Overall, we want to ensure an experiment that captures how RNNs learn time series structures, not why they have known instabilities. We then evaluate the performance (forecast accuracy) of the RNNs using mean squared error (MSE) and mean average percentage error (MAPE) on the standardized data to allow consistent comparisons of the outcomes. In terms of the computational resources, we use TensorFlow 2.1 on a single GTX 970 GPU for model training and activation layer extraction.   

\subsubsection{Time Series Process Results}
We now report the results of a series of simulations where a total of 50 runs are conducted for a given time series process. For each run, we train an RNN from scratch using a specified time series, calculate both the distance correlation values described in Section \ref{sec: RNN_limits}, and the corresponding MSE and MAPE values. Note that for all the time series plots, the values are de-standardized to their original scales in order to view the actual results.  

\begin{figure}[!ht]
	\centering
		\includegraphics[width=\textwidth]{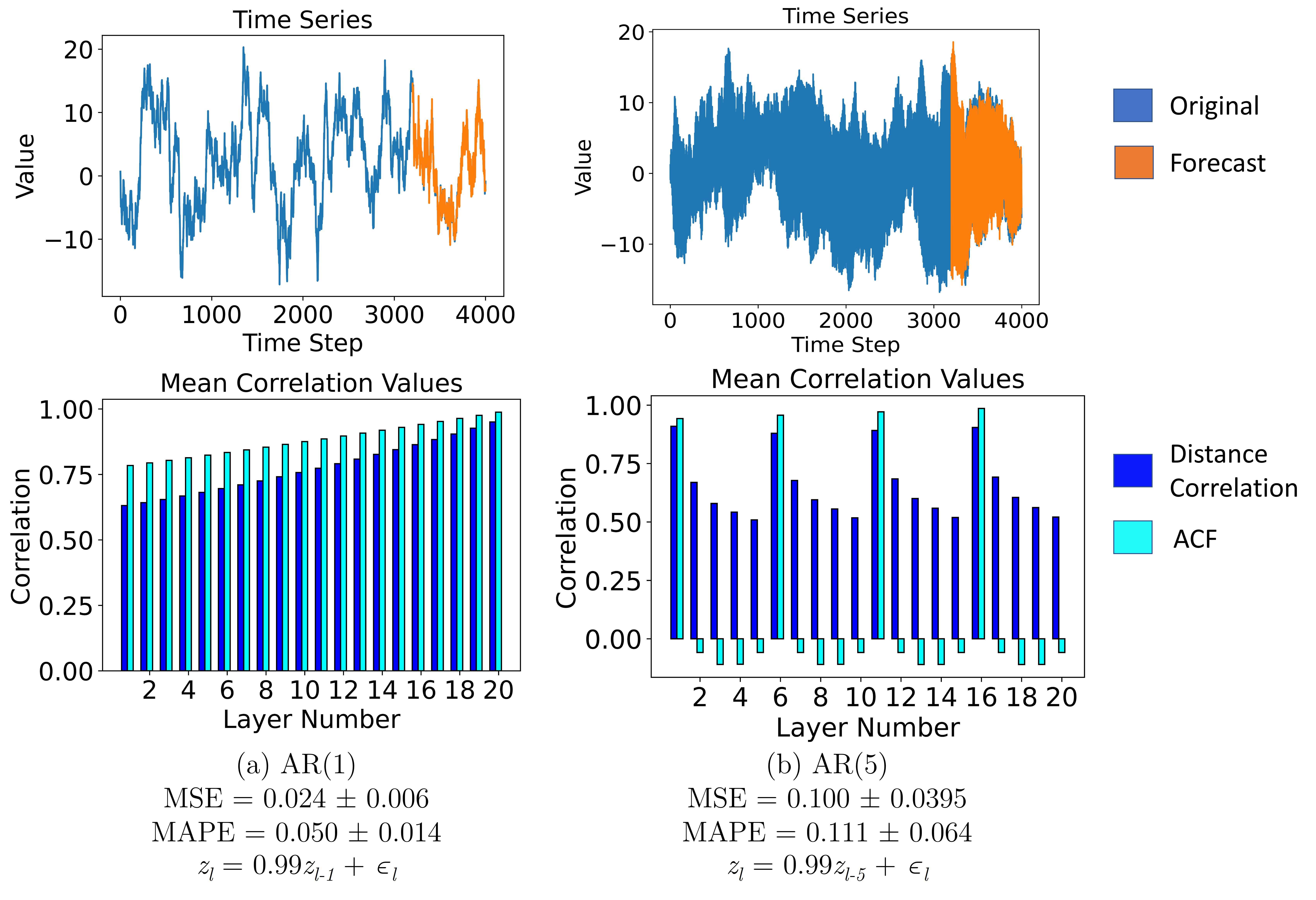}
	\caption{AR time series plots with RNN forecasts of the test set (top row). The mean values of ACF and the distance correlations between the outputs of activation layers and the ground truth horizon values are shown in the bar plots for 50 runs (bottom row). For the AR(1) process, we observe high correlation values for both metrics and a gradual increase as the layer number increases. For the AR(5) process, we see layers 1, 6, 11, and 16 with high correlations, but they cyclically diminish for every 5 layers. This aligns with high ACF values whose corresponding lags occur at 20, 15, 10, and 5. Both the de-standardized time series plots have well fitting forecasts from the RNN, which corresponds to the relatively low MSE and MAPE scores.}
	\label{fig:AR_1_5}
\end{figure}

\begin{figure}[!ht]
	\centering
		\includegraphics[width=\textwidth]{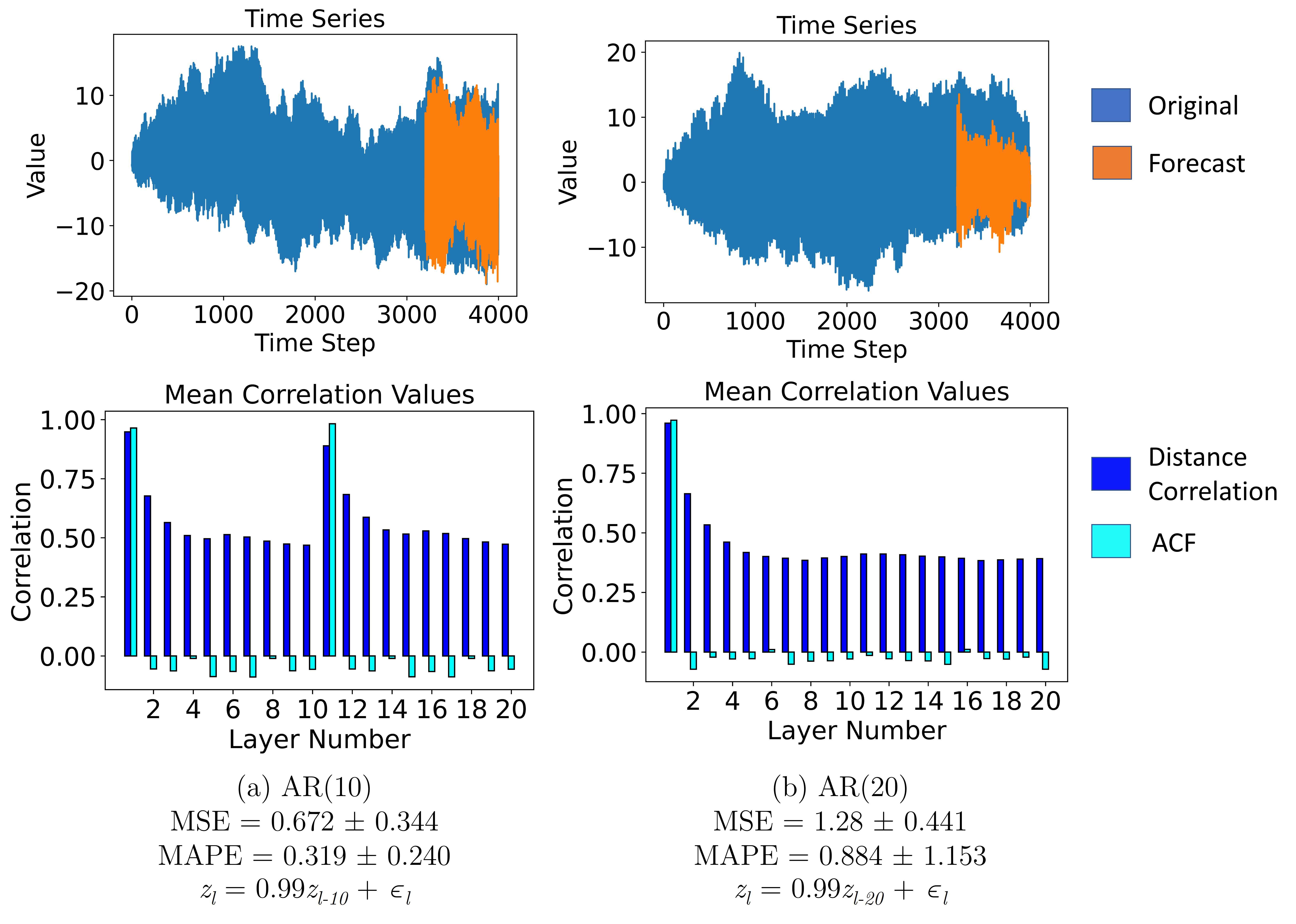}
	\caption{AR time series and mean correlation plots for 50 simulation runs. For AR(10), we see high distance correlation values at layers 1 and 11 with diminishing correlation values from layers 12 to 20. The high correlation values align with the ACF lags at layers 10 and 20. For AR(20), we only see high correlation at layer 1 (or lag 20), before a similar dissipation in values occur from layers 2 to 20. This decrease likely occurs due to the RNN losing some memory of the previous important inputs. We also see an increase in the MSE and MAPE values for both the processes, as compared to the AR(1) and AR(5) processes,
 with worse fits of the de-standardized forecasting plots.}
	\label{fig:AR_10_20}
\end{figure}

Beginning with auto-regressive time series, Figure \ref{fig:AR_1_5} shows two plots each for AR(1) and AR(5) time series processes: 1) A time series plot with the original and de-standardized forecast values by the trained RNN; and 2) the mean correlation values between the ground truth and each activation layer using both distance correlation and ACF metrics. Figure \ref{fig:AR_10_20} shows the same plots for AR(5) and AR(10) time series processes. In addition, the figures also display the corresponding MSE and MAPE values, and the coefficients for the time series processes.

Figure \ref{fig:AR_1_5}(a) shows that for an AR(1) process, the distance correlation values closely resemble the pattern produced by the ACF values for all the layers and lags. The similarity between ACF and distance correlation is likely due to the recursive nature of the AR(1) process, where all the time steps are recursively dependent on each other. There is a gradual increase in correlation as the layer numbers increase, which remain relatively high for all the layers (above 0.7). Noting that layer number 20 is the final activation layer prior to generating a forecast, it follows that a high distance correlation in this layer yields accurate forecasts. This is supported by the low MSE and MAPE values, and well fitting de-standardized time series forecasting plots. 

However, the corresponding plots for AR(5), AR(10) and AR(20) indicate that the forecasting accuracy of RNN decreases for increasing time series lag structures. Figure \ref{fig:AR_1_5}(b) shows the correctly identified AR(5) lag structure by displaying high correlation values at layers 1, 6, 11 and 16. Recalling the inverse relationship that lags have with the layer numbers, this aligns with the ACF values whose corresponding correlations are high at lags 20, 15, 10 and 5. This alignment, however, diminishes quickly as the distance correlation values decrease to approximately 0.5 at layer 20. This is further evident in Figure \ref{fig:AR_10_20}, where although the correct lags are identified by both distance correlation and ACF, distance correlation reduces to less than 0.4 for AR(10) and approximately 0.35 for AR(20) when it reaches layer 20; see Table \ref{table: All_Exp} for the exact correlation values at activation layer $T = 20$. This reduction in distance correlation with increasing layer numbers is synonymous to the RNN losing memory of the important previous inputs. As a result, the MSE and MAPE scores increase for the AR(10) and AR(20) time series, which is supported by the poor fitting de-standardized forecasting plots. 

\begin{figure}[!ht]
	\centering
		\includegraphics[width=\textwidth]{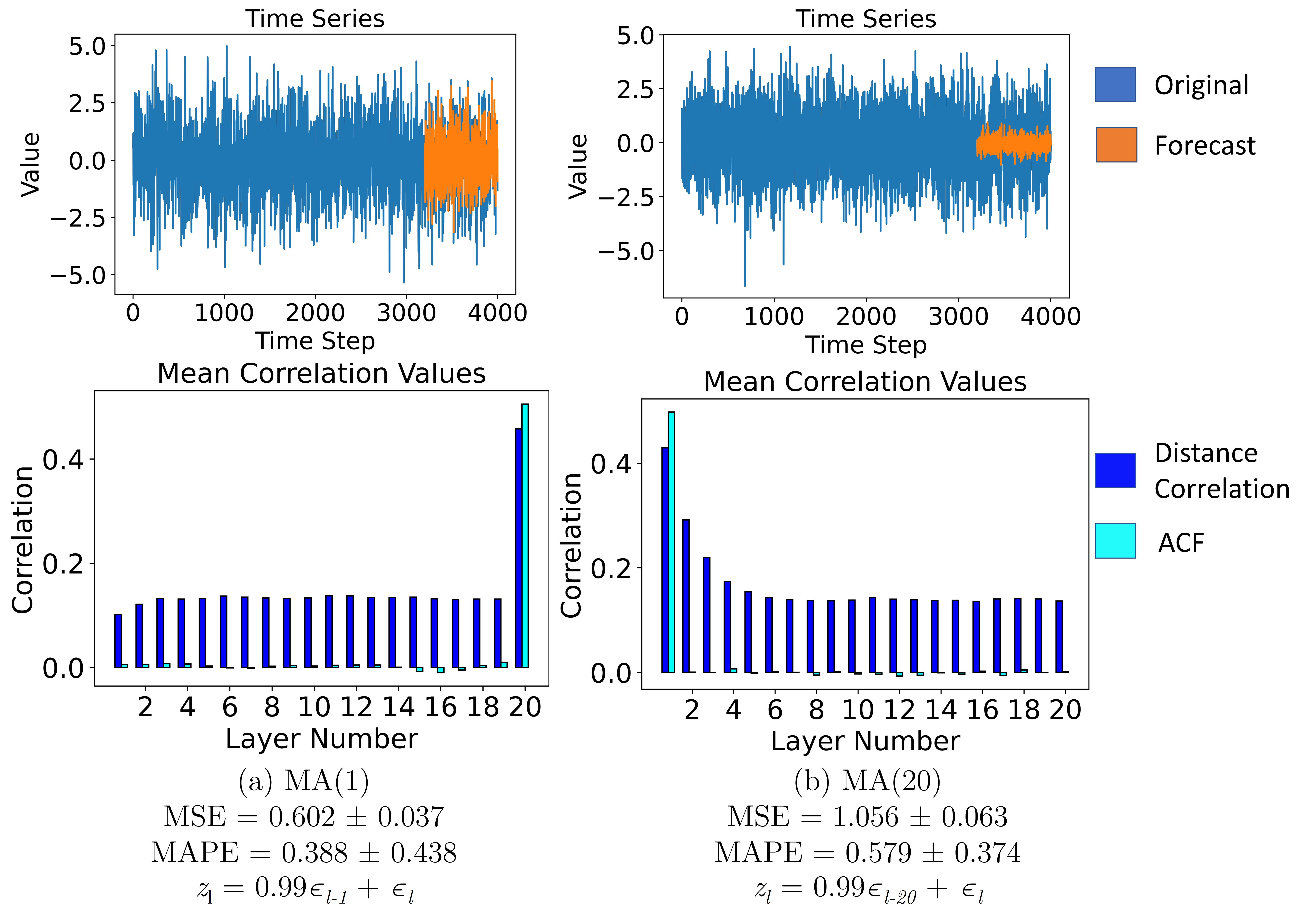}
	\caption{MA time series plots with RNN forecasts of the test set (top row). The mean values of ACF and distance correlation between the outputs of the activation layers and the ground truth horizon values are shown in the bar plots for 50 runs (bottom row). For both the MA(1) and MA(20) processes, we see that the lags are identified at layers 20 and 1 (lags 1 and 20), respectively, with the correlation values around 0.4 (0.5 for ACF). We also see for MA(20), distance correlation diminishes asymptotically and converges to a value of less than 0.1 at the final activation layer. This is analogous to the RNN losing memory of the important inputs, resulting in higher MSE and MAPE scores, and a worse fitting de-standardized forecasting plot.}
	\label{fig:MA_1_20}
\end{figure}

We extend these experiments to the moving average process by looking at MA(1) and MA(20) structures. Figure \ref{fig:MA_1_20} shows the same mean correlation and de-standardized time series forecasting plots. In Figure \ref{fig:MA_1_20}(a), the mean correlation plot displays the correct layer number (or lag structure) for MA(1); however, the value is relatively low (approximately 0.4). Although the distance correlation value is highest at layer number 20, the MSE and MAPE scores are larger than their AR(1) counterparts. The relatively low distance correlation value at the final activation layer combined with worse MSE and MAPE scores indicate that the RNN struggles to model the error lag structures well, even under ideal low lag structures.

Figure \ref{fig:MA_1_20}(b) shows the results for the MA(20) structure. Although the proper lag is identified at layer number 1, it is again relatively low with a value of approximately 0.4. However, since the lag structure is high (larger values of $p,q$) and far removed from the forecast horizon, we see an asymptotic decrease in correlation values near layer number 20, exhibiting values less than 0.1. The RNN appears to compensate for this lack of information being transmitted to the final layer by modeling values near the average (of zero), as opposed to the fluctuations that characterize the MA processes. This mean trend prediction is noticeable in the de-standardized time series forecasting plot where the amplitude is truncated and centered around zero. 

\begin{figure}[!ht]
	\centering
		\includegraphics[width=\textwidth]{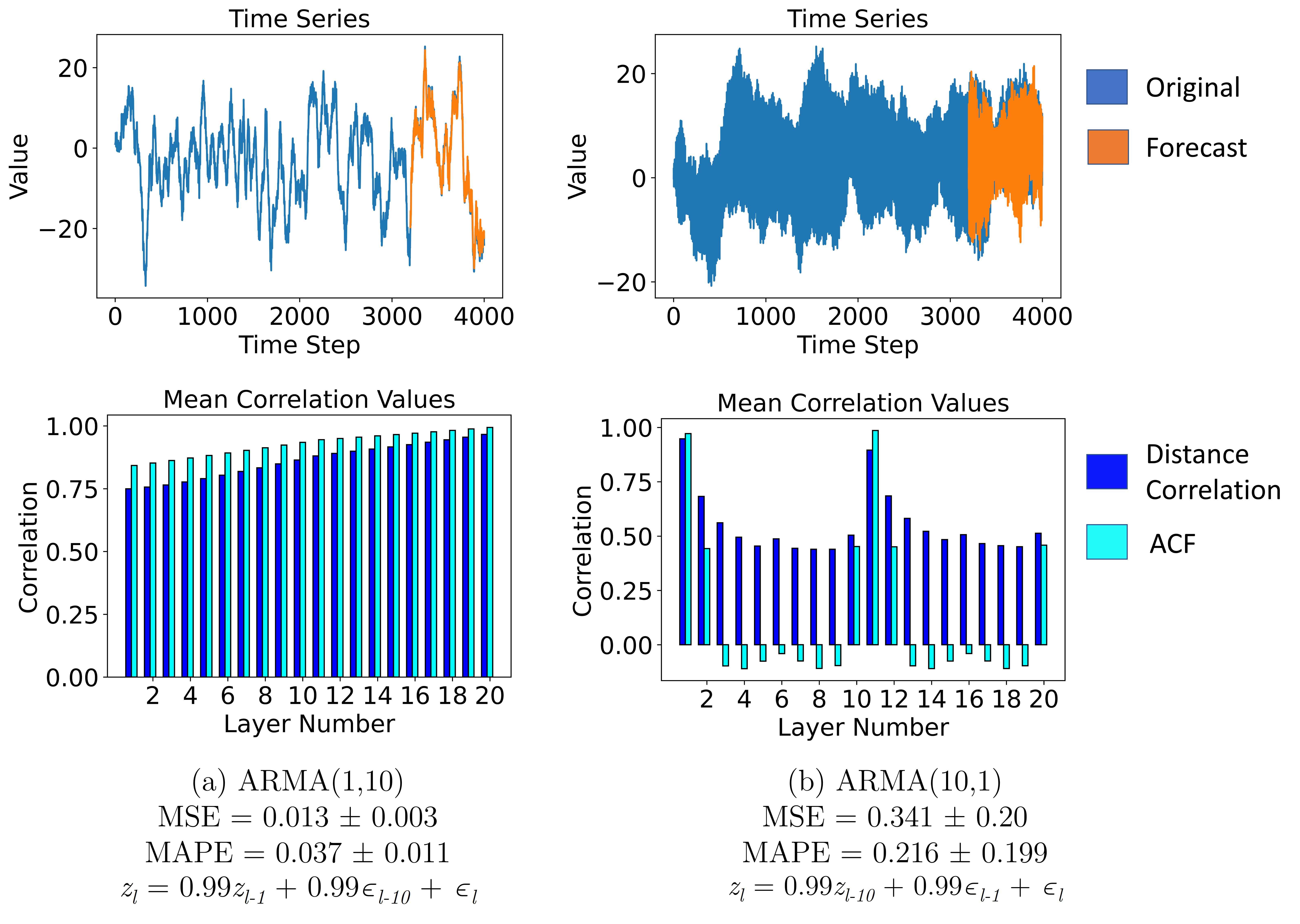}
	\caption{ARMA time series plots with RNN forecasts of the test set (top row). The mean values of ACF and the distance correlations between the outputs of the activation layers and the ground truth horizon values are shown in the bar plots for 50 runs (bottom row). For ARMA(1,10), high distance correlation values are exhibited for all the layers, which resemble the correlation plots of AR(1) and result in highly accurate forecasts. For ARMA(10,1), there is a notable lag structure similar to the AR(10) process. However, the decrease in RNN memory is less severe with a distance correlation value of 0.5 at layer 20, as compared to 0.4 for AR(10). This results in relatively well fitting de-standardized forecasting plots for both the ARMA processes.}
	\label{fig:ARMA_1_10_x2}
\end{figure}

We also look at the combined effects of AR($p$) and MA($q$) processes by modeling ARMA($p,q$) time series structures. Figure \ref{fig:ARMA_1_10_x2} shows the same set of plots for the ARMA(1,10) and ARMA(10,1) processes. Figure \ref{fig:ARMA_1_10_x2}(a) shows the results for ARMA(1,10), which exhibit a similar correlation behavior to that of AR(1). The distance correlation values are high for all the layers with the highest occurring at layer number 20. This results in highly accurate forecasts with low MSE and MAPE values, and well fitting de-standardized time series plots. Figure \ref{fig:ARMA_1_10_x2}(b) shows the results for ARMA(10,1), which exhibits similar correlation patterns to that of AR(10). However, the correlations do not decrease as much when approaching layer number 20 (0.5 instead of 0.4). The inclusion of the MA terms, therefore, seems to provide additional pertinent information on the underlying time series characteristics to the RNN. This slightly higher correlation values at the final layer leads to lower MSE and MAPE scores for both the ARMA processes as compared to their AR counterparts.

\begin{figure}[!ht]
	\centering
		\includegraphics[width=\textwidth]{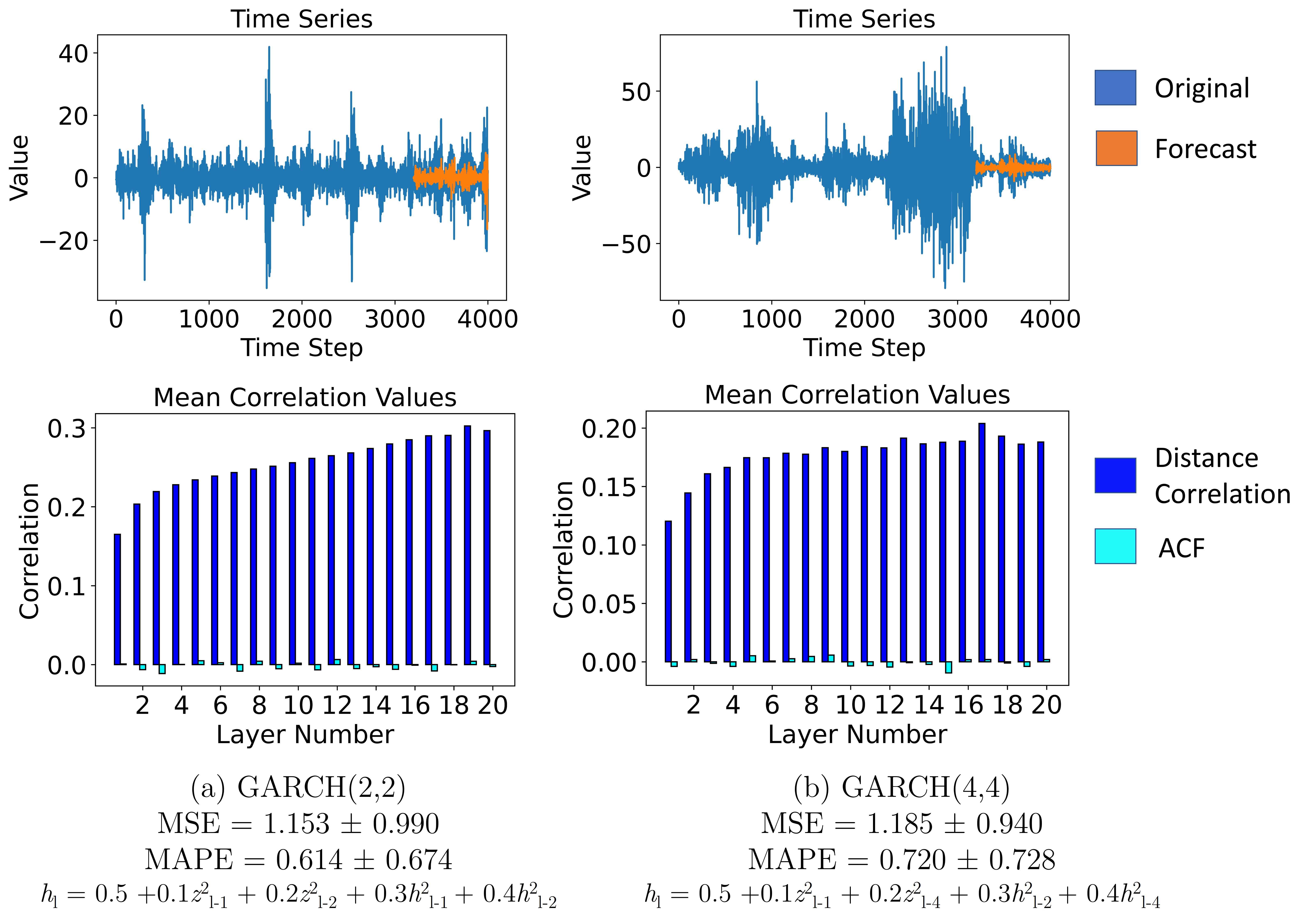}
	\caption{Time series plots with RNN forecasts of the test set (top row). The mean values of distance correlation and ACF between all the activation layer outputs and the ground truth horizon values are shown in a bar plot for 50 runs (bottom row). For both the GARCH(2,2) and GARCH(4,4) processes, we see that the distance correlation and ACF values are low; in fact, they are effectively zero with ACF for all the layer numbers. We also see that both the time series plots have poor RNN fitting forecasts, which corresponds to the relatively high MSE and MAPE scores. In particular, the RNN forecasts do not capture the spikes in variance well.}
	\label{fig:GARCH_22_44}
\end{figure}

We also explore how RNNs internally learn time series processes exhibiting heteroscedasticity with GARCH processes. We report the same time series forecasts and mean value correlation plots in Figure \ref{fig:GARCH_22_44}, where we note that the distance correlation values are quite low (less than 0.25) for all the layers. The low correlation values indicate that none of the RNN activation layers make strong contributions to learning the ground truth. This is corroborated by the poor fitting de-standardized forecasting plots for both the GARCH(2,2) and GARCH(4,4) processes. In essence, the RNN activation layers are not able to effectively model the heteroscedasticity and volatility present in the GARCH process, at least without additional pre-processing.

\begin{figure}[!h]
	\centering
		\includegraphics[width=\textwidth]{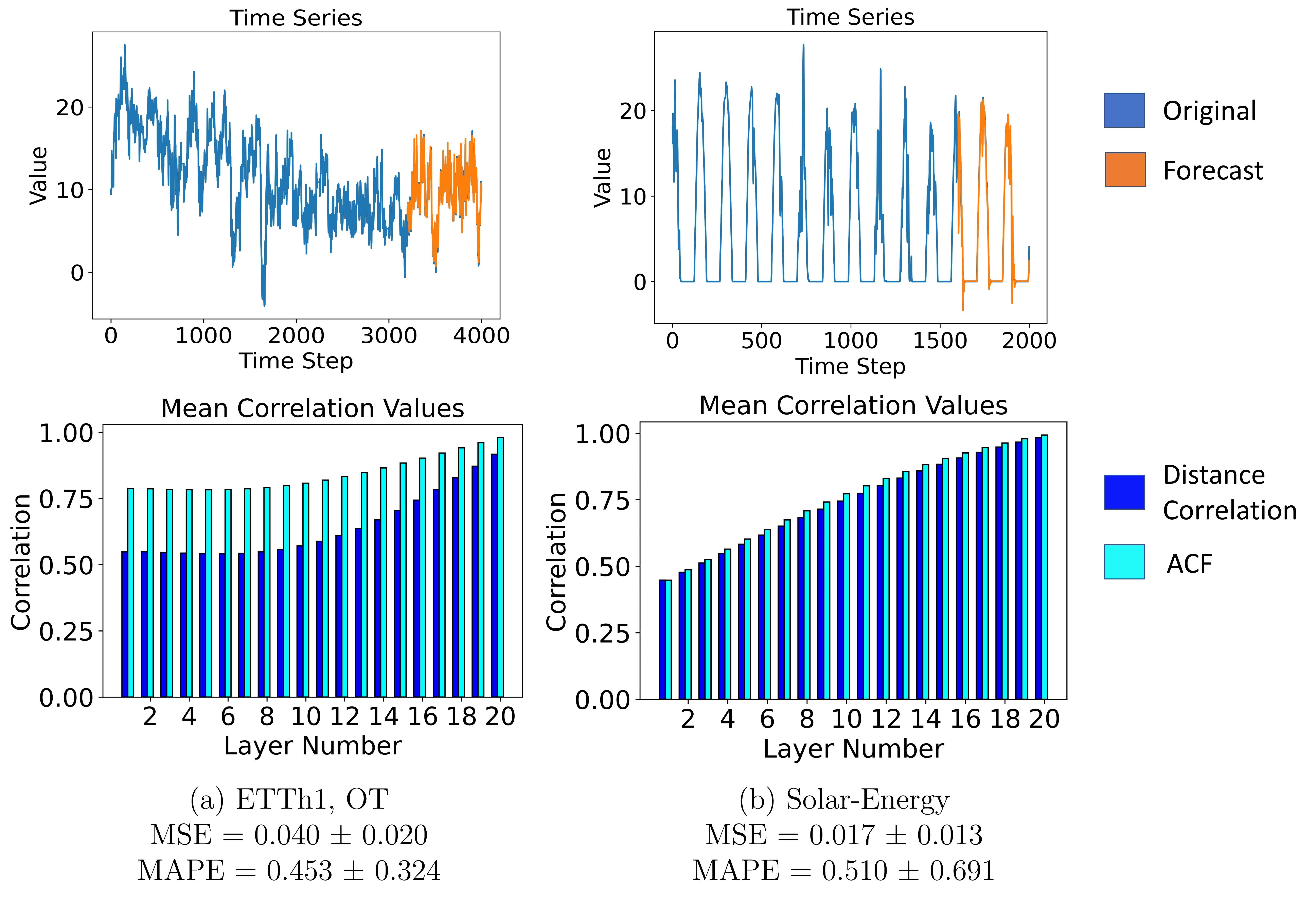}
	\caption{Time series plots with RNN forecasts of the test set (top row). The mean values of distance correlation and ACF between all the activation layer outputs and the ground truth horizon values are shown in a bar plot for 50 runs (bottom row). For both the ETTh1, OT and Solar-Energy data, we see that the distance correlation and ACF values gradually increase until layer 20 is reached. This results in  well fitting de-standardized forecasts, which correspond to the relatively low MSE and MAPE scores.} 
	\label{fig:Real_TS}
\end{figure}

\begin{figure}[!h]
	\centering
		\includegraphics[width=0.6\textwidth]{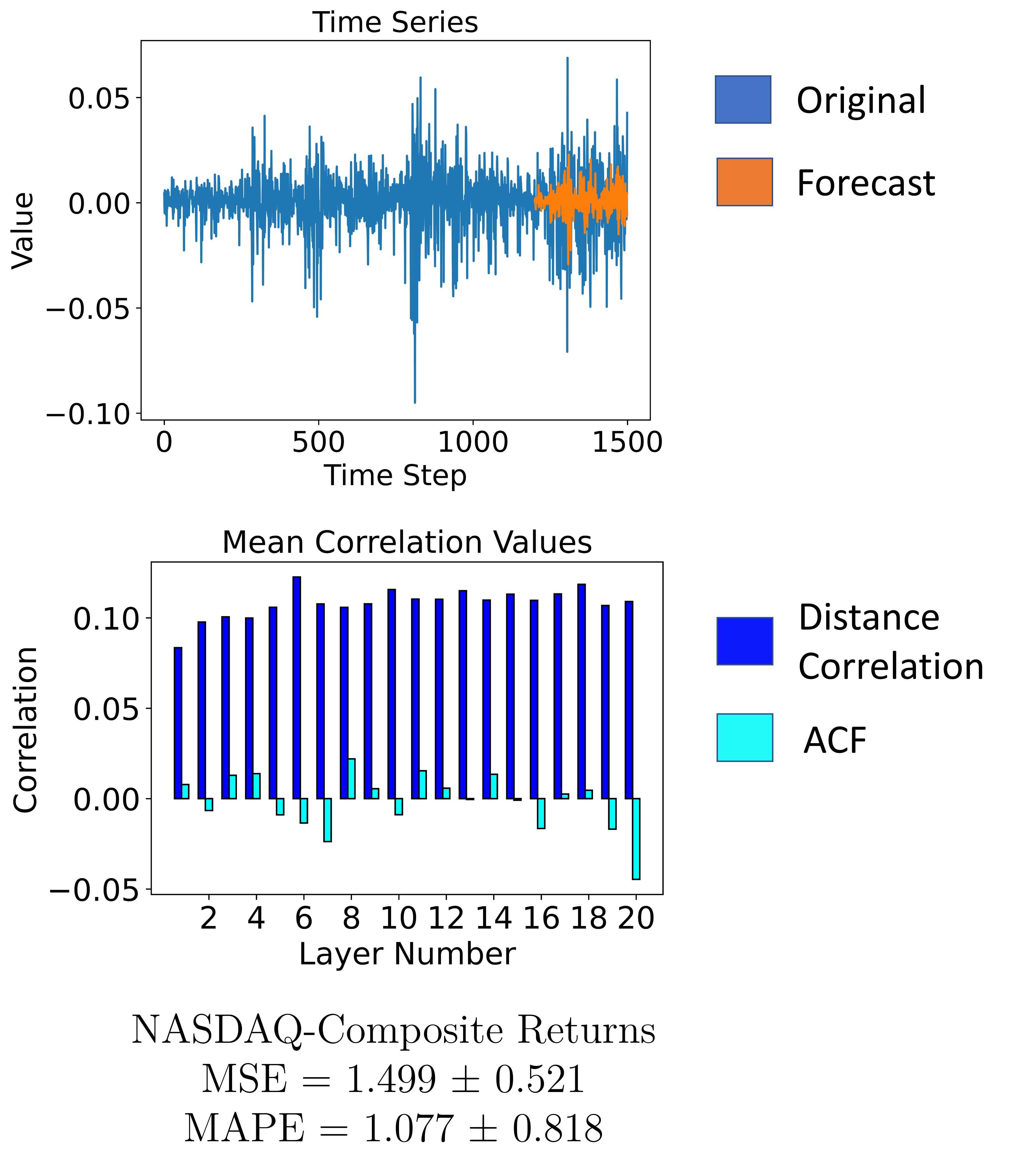}
	\caption{Time series plots with de-standardized RNN forecasts of the test set (top). The mean values of distance correlation and ACF between all the activation layer outputs and the ground truth horizon values are shown in a bar plot for 50 runs (bottom). For the NASDAQ Composite returns, we see that the distance correlation and ACF values are low; in fact, the ACF values are close to zero for all the layers. This leads to poor fitting forecasts, which correspond to the relatively high MSE and MAPE scores. This example is similar to the outcomes of the GARCH process in Figure \ref{fig:GARCH_22_44}.}
	\label{fig:NASDAQ}
\end{figure}

Last, we employ our evaluation method on the real world datasets described in Section \ref{sec: Gen_TS}. We generate the same correlation plots and de-standardized forecast plots for the ETTh1, OT and the Solar-Energy datasets in Figure \ref{fig:Real_TS}. Figure \ref{fig:Real_TS}(a) shows that  layers 1-10 have approximately constant correlation values of around 0.5 before gradually increasing to 0.8 at layer 20. This corresponds to having well fitting forecasts and lower MSE and MAPE scores. Similar observations are found in Figure \ref{fig:Real_TS}(b), where the  correlation values increase at a constant rate until a value of 0.85 at activation layer 20. As seen for the AR and ARMA processes, high correlation values at the final layer lead to fairly accurate forecasts.

However, the results for the NASDAQ Composite data in Figure \ref{fig:NASDAQ} are quite different. The correlation plots show that none of the activation layers recognize the underlying process exhibited by the financial data. All the correlation values are near 0.1 and there is no evidence that the activation layers learn the outputs well, leading to poor forecasting results. In fact, the results are similar to those for the GARCH process. In summary, these real world examples provide additional validity to the fact that distance correlation can effectively examine the behavior, especially, learning capacity, of RNN forceasting models for any given times series.

\subsubsection{Temporal information loss in RNNs}

We recall the studies that used distance correlation as a metric of the information stored in neural network activation layers \cite{zhen2022versatile, vepakomma2020nopeek}, and apply the same principle in our study. From our experiments, we observe 
that the distance correlation values change as the time series inputs move from one activation layer to the next, analogous to a measurable change in information content across the RNN. Table \ref{table: All_Exp} summarizes the results for all the time series forecasting experiments, and report the metrics that encode this RNN information flow. The key metric 
is the percentage change between the max distance correlation 
at any layer 
and the distance correlation at the final layer. 
This is calculated by $(\hat{R}(\textbf{A}_{t}^{(P)}, \textbf{Y})_{max} - \hat{R}(\textbf{A}_{T}^{(P)}, \textbf{Y})) / \hat{R}(\textbf{A}_{t}^{(P)}, \textbf{Y})_{max}$.  It summarizes how much information is lost by the time the data reaches the final activation layer before the RNN produces a forecasting value. 

From Table \ref{table: All_Exp}, we see that for the higher AR lag structures, the RNN is able to identify the proper lags. However, higher lag structure AR processes lose a larger percentage of the information as they reach the final activation layer. For the MA lag structures, the proper lags are also identified, although they exhibit smaller distance correlation values. Nonetheless, the higher lag structures still lose more information as they reach the final activation layer. This general loss of memory in the final activation layer leads to higher MSE and MAPE scores. This is also exemplified by comparing AR(10) to ARMA(10,1), where they have similar distance correlation patterns, as seen in Figures \ref{fig:AR_10_20}(a) and \ref{fig:ARMA_1_10_x2}(b). However, ARMA(10,1) loses 48\% of the input memory as compared to 59\% for AR(1), resulting in lower MSE and MAPE values. The exception to this behavior occurs for the GARCH and NASDAQ time series, where no RNN activation layer learns the ground truth values well, leading to poor overall results.

These experiments and outcomes were replicated for different RNN hyperparameters to show that this behavior is consistent despite changes in the modeling parameters. See Appendix A for a summary of the additional results that include changes to the number of hidden units, learning rate, and dropout rate of the final RNN layer.

\begin{table}[!htbp]
\centering
\caption{MSE, MAPE, and distance correlation results for all the experiments}
\resizebox{\textwidth}{!}{
\begin{tabular}{||p{0.16\textwidth} | p{0.10\textwidth} | p{0.12\textwidth} | p{0.14\textwidth} | p{0.14\textwidth} | p{0.14\textwidth}||} 
 \hline
 Time Series & MSE & MAPE & $\hat{R}(\textbf{A}_{t}^{(P)}, \textbf{Y})$ (max) & $\hat{R}(\textbf{A}_{T}^{(P)}, \textbf{Y})$ & Change (in \%) \\ [0.5ex] 
 \hline\hline
 AR(1) & 0.025 $\pm$ 0.006 & 0.050 $\pm$ 0.014 & 0.927 & 0.927 &  0 \\ 
 \hline
 AR(5) & 0.100 $\pm$ 0.039 & 0.111 $\pm$ 0.064 & 0.916 &  0.436 & 52 \\ 
 \hline
 AR(10) & 0.672 $\pm$ 0.344 & 0.319 $\pm$ 0.240 & 0.949 & 0.391 & 59 \\ 
 \hline
 AR(20) & 1.281 $\pm$ 0.441 & 0.884 $\pm$ 1.154 & 0.964 & 0.356 & 63 \\ 
 \hline
 MA(1) & 0.602 $\pm$ 0.037 & 0.388 $\pm$ 0.438 & 0.418 & 0.418 & 0 \\ 
 \hline
 MA(5) & 0.783 $\pm$ 0.057 & 0.525 $\pm$ 0.603 & 0.427 & 0.131 & 69 \\ 
 \hline
 MA(10) & 1.008 $\pm$ 0.081 & 0.512 $\pm$ 0.235 & 0.422 & 0.098 & 77 \\ 
 \hline
 MA(20) & 1.056 $\pm$ 0.063 & 0.579 $\pm$ 0.374 & 0.435 & 0.093 & 79 \\
 \hline
 ARMA(1,1) & 0.009 $\pm$ 0.003 & 0.035 $\pm$ 0.012 & 0.935 & 0.935 & 0 \\ 
 \hline
 ARMA(1,10) & 0.013 $\pm$ 0.003 & 0.037 $\pm$ 0.011 & 0.951 & 0.951 & 0 \\ 
 \hline
 ARMA(10,1) & 0.341 $\pm$ 0.195 & 0.216 $\pm$ 0.199 & 0.947 & 0.496 & 48 \\ 
 \hline
 GARCH(2,2) & 1.037 $\pm$ 0.830 & 0.614 $\pm$ 0.674 & 0.264 & 0.259 & 2 \\ 
 \hline
 GARCH(4,4) & 1.080 $\pm$ 0.751 & 0.720 $\pm$ 0.728 & 0.215 & 0.202 & 6 \\ 
 \hline
 ETTh1, OT & 0.040 $\pm$ 0.020 & 0.453 $\pm$ 0.324 & 0.923 & 0.923 & 0 \\ 
 \hline
 Solar-Energy & 0.017 $\pm$ 0.013 & 0.510 $\pm$ 0.691 & 0.983 & 0.983 & 2 \\ 
 \hline
 NASDAQ & 1.499 $\pm$ 0.521 & 1.077 $\pm$ 0.818 & 0.123 & 0.114 & 7 \\ 
 \hline
\end{tabular}\label{table: All_Exp}
}
\begin{minipage}{\textwidth}
\vspace{2mm}
\tiny Notes: MSE and MAPE scores are calculated for 50 simulation runs. $\hat{R}(\textbf{A}_{t}^{(P)}, \textbf{Y})$ (max) represents the max distance correlation for any layer number $t$. $\hat{R}(\textbf{A}_{T}^{(P)}, \textbf{Y})$ is the distance correlation value at layer number $T = 20$. The percent change from the peak distance correlation values in any activation layer to the final layer measures how much information is lost during RNN training. In general, we see that larger lag structures tend to lose more information during training, leading to higher MSE and MAPE scores.
\end{minipage}
\end{table}

\subsection{Visualizing differences in time series RNN models}

\begin{figure}
	\centering
		\includegraphics[width=0.5\textwidth]{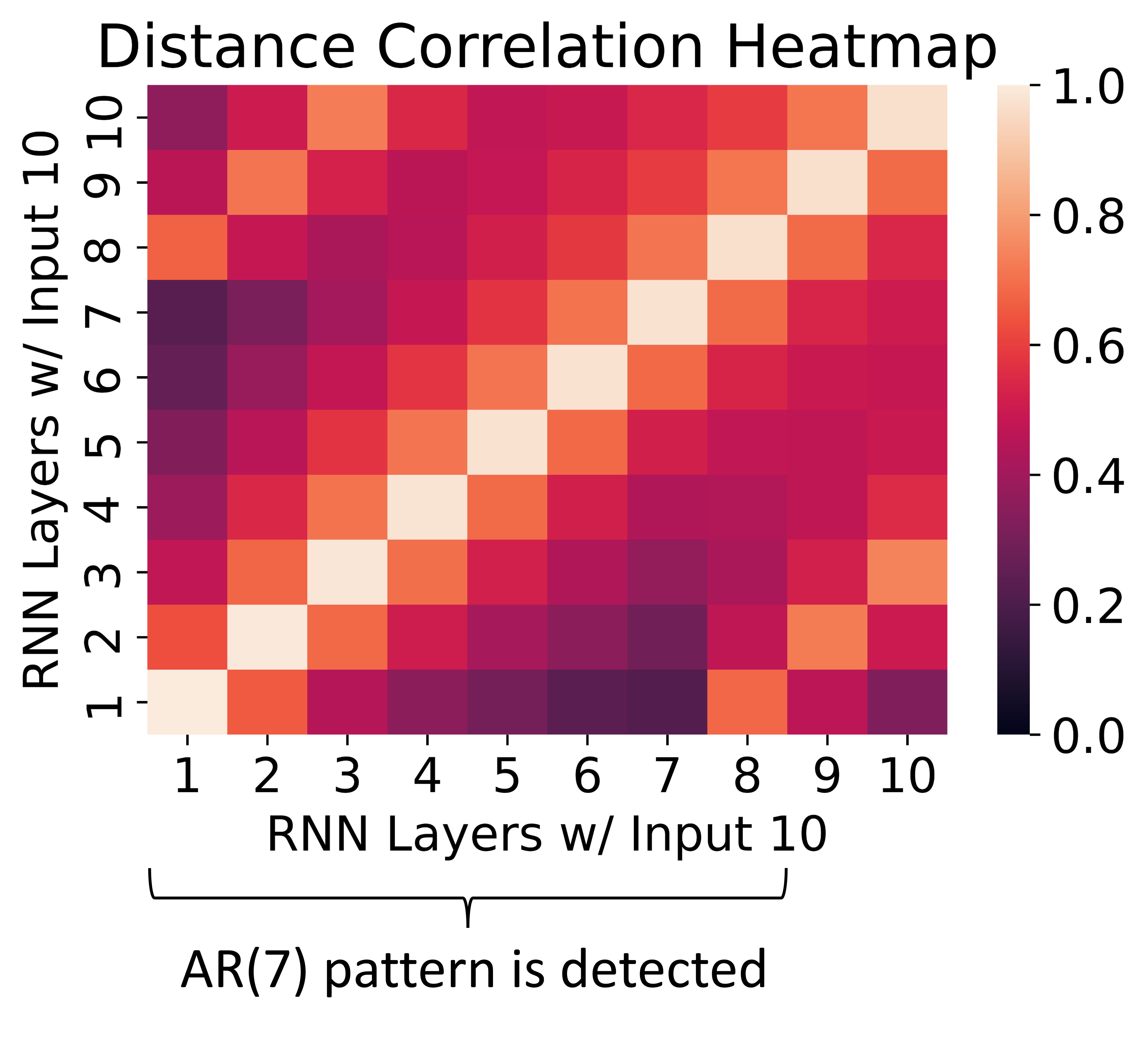}
	\caption{Distance correlation heatmap between RNN with 10 inputs and itself. The RNN was trained on an AR(7) process with the following governing equation: $z_{l} = 0.8z_{l-7} + \epsilon_{l}$. We observe there is an exact symmetry along the diagonal, where the distance correlation between each RNN layer and itself is 1. We also see that the AR(7) pattern is detected if we follow the progression of values from layer 1 on the y-axis to layer 8 of the x-axis.}
	\label{fig:Heat_same}
\end{figure}

As we try to gain a better understanding of RNN capabilities with time series forecasting, we leverage distance correlation to generate heatmaps as a visual way of investigating the activation layers; similar to the comparison of computer vision models \cite{zhen2022versatile}. We motivate this visualization method by considering which RNN hyperparameters lead to accurate forecasts. These include input size, activation functions, the number of hidden units, and even the RNN output configurations (i.e., many-to-one or many-to-many). For example, Figure \ref{fig:Heat_same} shows a distance correlation heatmap between a trained RNN with 10 inputs (activation layers) and itself, where we see a complete symmetry in the heatmap. This baseline heatmap can help us visualize the similarities between the activation layers at different time steps and reveal characteristics such as the lag structure. We expand such comparisons with some of the hyperparameters mentioned above. 

\begin{figure}[!ht]
	\centering
		\includegraphics[width=\textwidth]{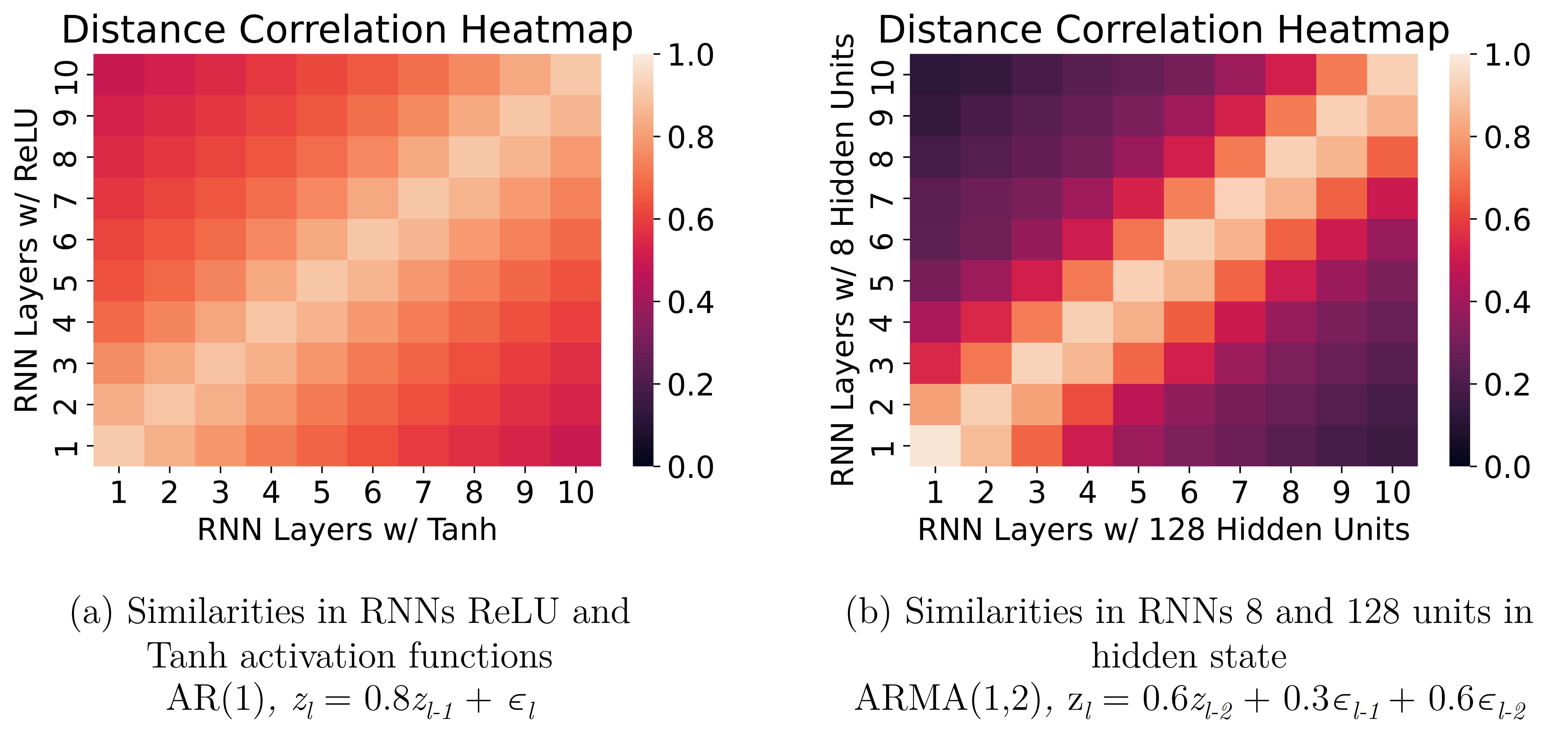}
	\caption{Visualization of the similarities of different RNNs in time series modeling using distance correlation heatmaps. (a) 
 The symmetry in the heatmap suggests that using either activation function yields similar outputs at each activation layer. (b) 
 The heatmap also suggests that both the networks arrive at similar activation layer outputs, despite the differences in the hidden unit size of the activation layers.}
	\label{fig:Sim_Network}
\end{figure}

Distance correlation heatmaps can show whether two networks with different hyperparameters are similar at each activation layer. Two examples of this are displayed in Figure \ref{fig:Sim_Network}, where we test networks with different activation functions and number of hidden units. In the case of Figure \ref{fig:Sim_Network}(a), the symmetric nature of the heatmap suggests that using either ReLU or Tanh activation function makes minimal difference in its evaluation of an AR(1) process. Extending this to Figure \ref{fig:Sim_Network}(b), we see a similar symmetry in the heatmap for networks having 8 and 128 hidden units in the activation layer for an ARMA(1,2) process. Note that all the networks were allowed to train until convergence, which exceeded 100 epochs for the RNN with 8 hidden units. Both these results show that the different hyperparameter choices do not have any noticeable effect for a given time series process, provided the computational cost is not an issue. 

\begin{figure}[!ht]
	\centering
		\includegraphics[width=0.9\textwidth]{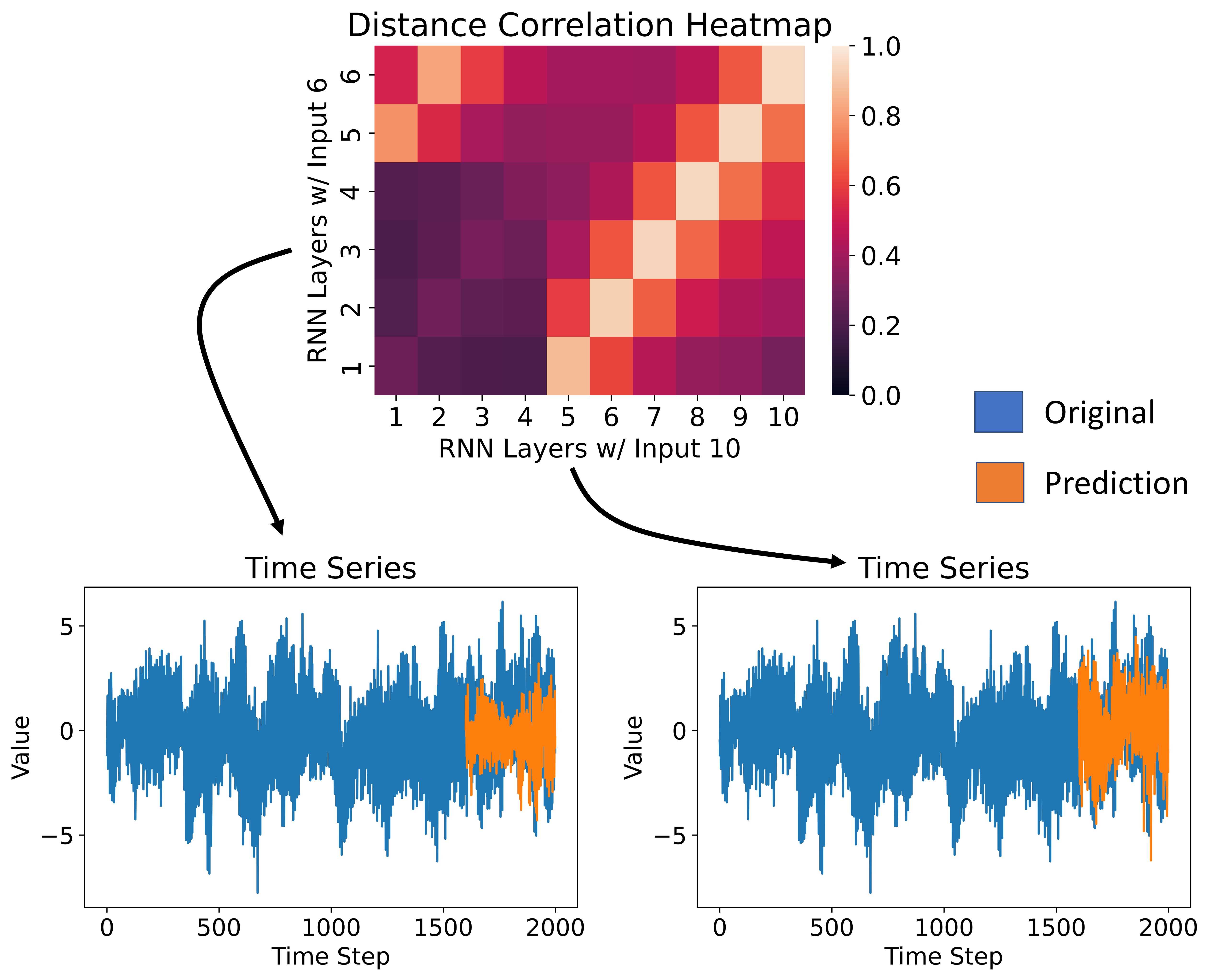}
	\caption{Distance correlation heatmap between RNN with 6 inputs and 10 inputs. This example invokes a scenario where the input size is smaller than the largest lag in the AR process. The RNNs were trained under an AR(8) process with the following governing equation: $z_{l} = 0.8z_{l-8} + \epsilon_{t}$. The 6 input and 10 input RNNs received an MSE = 1.18 $\pm$ 0.056 (MAPE = 0.510 $\pm$ 0.0816) and 0.496 $\pm$ 0.024 (MAPE = 0.272 $\pm$ 0.226), respectively, after 50 simulation runs. RNN (6 input) activations layers 1-6 and RNN (10 input) activation layers 5-10 are noticeably highly correlated. This correlation suggests that the smaller network learns the final activation layer outputs of sufficiently sized networks. However, without access to the 8th lagged time step from the AR(8) process, the smaller network is missing crucial information that prevents accurate forecasts. This is evident from the worse fitting forecasts of the time series plot and the corresponding MSE and MAPE scores.}
	\label{fig:6_10_Net}
\end{figure}

Another crucial choice with RNN time series forecasting is the proper window input size. While some heuristics can be applied in choosing this parameter, we can also use distance correlation heatmaps to understand the impact. We exemplify this with a heatmap that compares an RNN with 6 inputs and 10 inputs under an AR(8) process which is shown in Figure \ref{fig:6_10_Net}. In this scenario, the smaller RNN shows that it can learn similar activation outputs to the larger 10 input RNN as shown by the similarities in the diagonal elements between activation layers 1-6 and 5-10, respectively. However, it seems that without having access to at least 8 inputs for the AR(8) process, suggests that the smaller network cannot adjust its learning weights via a backpropagation through time algorithm \cite{werbos1990backpropagation} to produce accurate forecasts. This is supported by the lack of good fit in the de-standardized time series forecast plot for the 6 input RNN and the better fit in the 10 input RNN. 

\begin{figure}[!ht]
	\centering
		\includegraphics[width=0.9\textwidth]{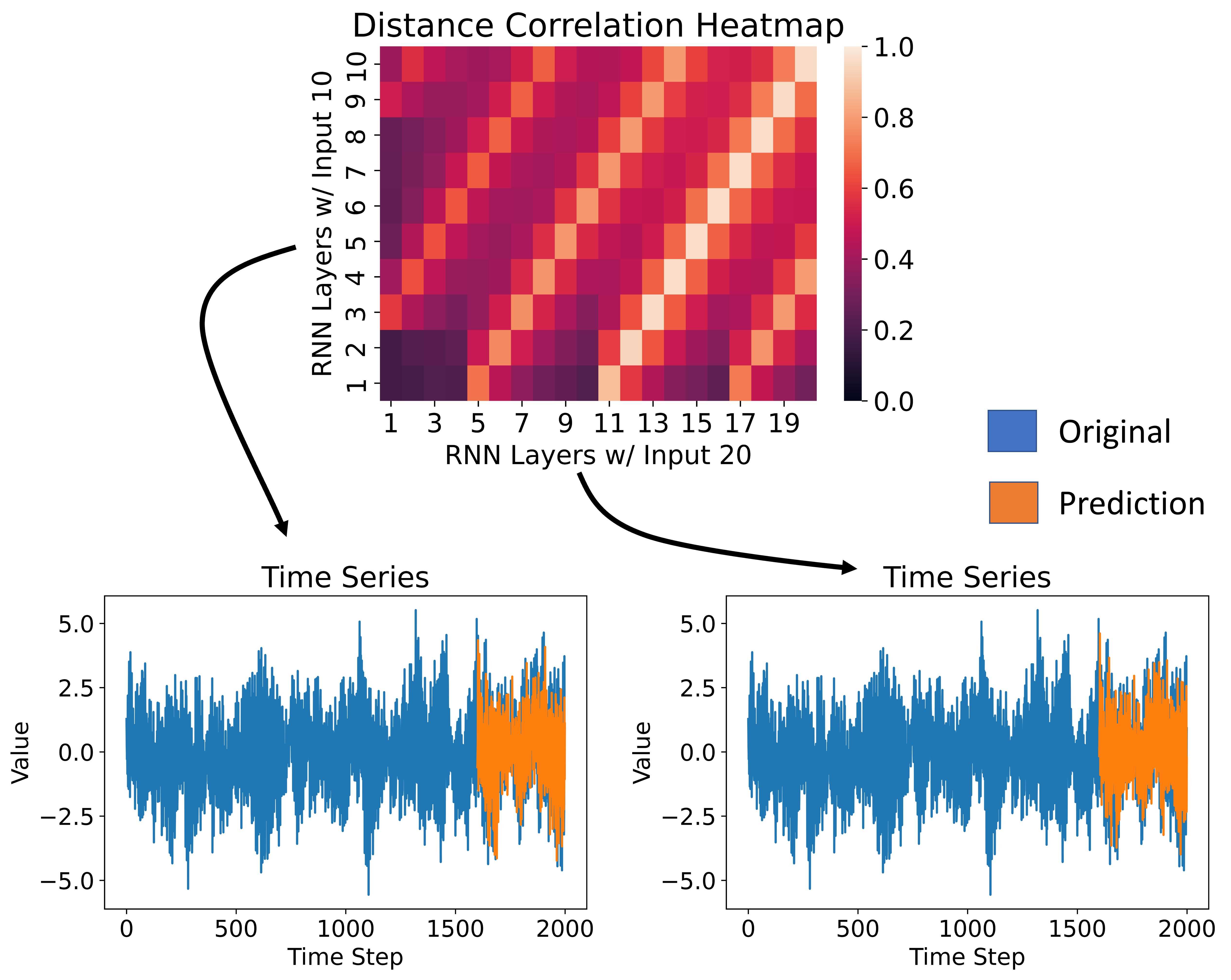}
	\caption{Distance correlation heatmap between RNNs with 10 inputs and 20 inputs. The RNNs were trained under an AR(6) process with the following governing equation: $z_{l} = 0.8z_{l-6} + \epsilon_{t}$. The 10 input and 20 input RNNs received an MSE of 0.501 $\pm$ 0.061 (MAPE = 0.278 $\pm$ 0.0790) and 0.490 $\pm$ 0.045 (MAPE = 0.293 $\pm$ 0.137), respectively, after 50 simulation runs. The heatmap shows diagonal streaks of similarities in both the networks, which are spaced apart by 6 activation layers, matching the AR(6) process. The 20 input RNN encounters the AR(6) lag structure at least thrice, as compared to twice in the 10 input RNN. This may be the source of why the former RNN yields a slightly lower MSE (and comparable MAPE) than the latter RNN. However, the corresponding time series forecasting plots indicate that both the RNNs learn the time series structure adequately as their input sizes are sufficiently large.}
	\label{fig:10_20_Net}
\end{figure}

We can then adopt a conservative strategy of 
choosing a window input size 
that is larger than necessary. Putting aside the known issues with long term dependency and exploding gradient problems in RNNs, we use distance correlation heatmaps to identify how the RNNs learn time series when the networks inputs are oversized. Figure \ref{fig:10_20_Net} simulates this scenario where both the RNN input sizes (10 and 20, respectively) are greater than the AR(6) process lag structure. We notice that the diagonal streaks of similarities are 6 activation layers apart, which confirms that both the RNNs can detect the 6 lag structure of the time series. Further, the 20 input RNN encounters the lag structure multiple times. This redundancy likely aids in the backpropagation through time algorithm, where the adjusted layer weights are updated to have more accurate forecasts. 

\section{Discussion}

We use distance correlation to show that the RNN activation layers can identify lag structures but have issues in transmitting that information to the final activation layer. This loss in information seems to occur over a span of 5-6 activation layers before the distance correlation values converge. The critical implication 
is that the RNN forecasts of 
univariate time series processes with large lag structures are likely to be poor 
since they are subject to more information loss. This can affect design decisions such as the sampling rate. For example, 
a sampling rate of hours, as compared to days, may subject the RNN to larger lag structures, even though a higher resolution input sequence is obtained. 
Conversely, lower lag structures make RNNs a preferred model with 
fast forecast results as they are relatively computationally inexpensive. These results are supported by our evaluation on the ETTh1, OT and Solar-Energy datasets, where we observe high correlation values at the final activation layers. This yields favorable forecasts and indicates that the RNNs have adequate learning capacity for single-step forecasting of such data.

Additionally, low distance correlation values of the activation layers in MA processes show that the error lags are difficult to model for RNNs. This lower accuracy is exacerbated by the information loss of RNNs. We also observe that the low distance correlation values in all the activation layers for GARCH processes lead to poor performance. Thus, an RNN model of any heteroskedastic time series 
is not likely to perform well. Relating this to real world data, we observe that the NASDAQ Composite returns look similar to the GARCH process results. None of the activation layer correlations are high, leading to poor fitting forecasts. In these instances, we find that the low correlations in the activation layers relate to the lack of the RNN's learning capacity given this form of time series data.

These results can also help the practitioners with \textit{a priori} assessments of the suitability of RNN forecasts given the characteristics of their time series data. 
For example, ACF plots, indicating whether the lag structures are high, can determine whether the corresponding time series processes are suitable for RNN forecasting. 
The presence of AR, MA, ARMA, and GARCH processes are also indicated in the ACF plots, providing cues on whether RNNs would be suitable for the corresponding forecast problems. 
This knowledge is valuable in reducing unnecessary and time-consuming model building and exploration steps. 
Coupling these observations with the 
hyperparameter-based comparisons provided by the distance correlation heatmaps, analysts are now better equipped 
to both interpret and explain RNN modeling of time series.

We acknowledge some ways to improve and expand our findings. First, we primarily consider well-established time series processes. This is done intentionally to create a controlled experiment, where the inputs and outputs of the time series processes are well defined. An expansion of the scope of our experiments to real world datasets, may provide a more practical context for our evaluation tool. Further, we only consider the most basic form of time series forecasting, which is univariate, single-step prediction. Single-step predictions can make our plotted forecasting results look quite favorable, but they should not be confused with the much more challenging task of forecasts with larger prediction horizons. Multivariate and multiple horizon forecasting are expected to reveal more insights into the effectiveness of RNNs, and we believe that distance correlation is powerful and generic enough to be adapted to such problems.

We also recognize that our distance correlation-based analysis is done with only Elman RNNs. However, we contend that it can be 
expanded to other widely-used RNN architectures, such as 
Long-Short Term Memory (LSTM) and Gated Recurrent Unit (GRU) models, which have more complex cell operations to alleviate some of the common pitfalls of RNNs. It may be particularly interesting to investigate how the information is partitioned among 
the regulatory gates and additional recurrent cell states of LSTMs.
This analysis tool can also be extended to many other deep learning networks that attempt to address time series forecasting. This includes observing how a time series input evolves through every major component in 
a transformer or another hybrid architecture. Overall, we feel that distance correlation is a flexible metric that can potentially unlock the unexplained nature of deep learning models for time series forecasting tasks. 

\section{Conclusions}

In this paper, 
we develop a distance correlation framework 
to study the effectiveness of RNNs for time series forecasting.
Specifically, we leverage the versatility of distance correlation to track the outputs of the RNN activation layers and examine how well they learn specific time series processes. Through both synthetic and real world data experiments, we find that the activation layers detect time series lag structures well, but tend to lose this information over a sequence of five-six layers. This affects the forecast accuracy of series with large lag structures. Further, the activation layers have difficulty in modeling the error lag terms for both moving average and heteroscedastic processes. Last, our distance correlation heatmap comparisons reveal that certain network hyperparameters, such as the number of hidden units and activation function, matter less as compared to the input window size of an RNN for accurate forecasts. We, therefore, believe our framework is a foundational step toward improving our understanding of the ability of deep learning models  
to handle time series forecasting tasks.


\appendix

\clearpage 

\section{Additional Experimental Results}\label{Add_Exp_Results}

The following tables report the time series forecasting results using RNN models that are slightly different to the model used in Table \ref{table: All_Exp}. These tables expand the RNN architecture to a larger number of input hidden units of 128 (Table \ref{table: HN_128}), a learning rate of 0.001 (Table \ref{table: LR_001}), and a dropout rate of 0.2 in the final layer (Table \ref{table: DO_02}). The purpose of these additional tables is to show consistency in the results from Section \ref{sec: RNN_limits}, even with a different set of hyperparameters. For example, each table shows that for increasing lag structures, MSE and MAPE scores grow, and the activation memory loss increases (shown by the change in \% column). It also shows that despite the changes to the hyperparameters, large MA lag structures and GARCH processes are not adequately modeled by an RNN. Given the agreement of these results, we contend that other changes to the RNN hyperparameters will lead to the same general outcomes.

\begin{table}[h!]
\caption{MSE, MAPE, and distance correlation results for all the experiments with 128 input hidden units}
\begin{tabular}{||p{0.16\textwidth} | p{0.12\textwidth} | p{0.12\textwidth} | p{0.15\textwidth} | p{0.15\textwidth} | p{0.15\textwidth}||} 
 \hline
 Time Series & MSE & MAPE & $\hat{R}(\textbf{A}_{t}^{(P)}, \textbf{Y})$ (max) & $\hat{R}(\textbf{A}_{T}^{(P)}, \textbf{Y})$ & Change (in \%) \\ [0.5ex] 
 \hline\hline
 AR(1) & 0.024 $\pm$ 0.005 & 0.058 $\pm$ 0.039 & 0.952 & 0.952 &  0 \\ 
 \hline
 AR(5) & 0.070 $\pm$ 0.029 & 0.105 $\pm$ 0.078 & 0.913 &  0.449 & 45 \\ 
 \hline
 AR(10) & 0.404 $\pm$ 0.197 & 0.275 $\pm$ 0.192 & 0.947 & 0.402 & 58  \\ 
 \hline
 AR(20) & 1.112 $\pm$ 0.333 & 0.809 $\pm$ 1.275 & 0.966 & 0.378 & 61  \\ 
 \hline
 MA(1) & 0.612 $\pm$ 0.034 & 0.303 $\pm$ 0.050 & 0.434 & 0.434 & 0  \\ 
 \hline
 MA(5) & 0.772 $\pm$ 0.050 & 0.434 $\pm$ 0.178 & 0.431 & 0.134 & 69  \\ 
 \hline
 MA(10) & 0.920 $\pm$ 0.059 & 0.424 $\pm$ 0.159 & 0.431 & 0.105 & 76  \\ 
 \hline
 MA(20) & 1.080 $\pm$ 0.064 & 0.572 $\pm$ 0.309 & 0.428 & 0.096 & 78  \\
 \hline
 ARMA(1,1) & 0.008 $\pm$ 0.002 & 0.035 $\pm$ 0.019 & 0.947 & 0.947 & 0  \\ 
 \hline
 ARMA(1,10) & 0.012 $\pm$ 0.003 & 0.040 $\pm$ 0.025 & 0.966 & 0.966 & 0 \\ 
 \hline
 ARMA(10,1) & 0.235 $\pm$ 0.137 & 0.251 $\pm$ 0.344 & 0.946 & 0.482 & 49  \\ 
 \hline
 GARCH(2,2) & 0.85 $\pm$ 0.674 & 0.651 $\pm$ 0.551 & 0.272 & 0.267 & 2  \\ 
 \hline
 GARCH(4,4) & 1.186 $\pm$ 1.081 & 1.406 $\pm$ 4.608 & 0.218 & 0.208 & 5  \\ 
 \hline
  ETTh1, OT & 0.038 $\pm$ 0.019 & 0.603 $\pm$ 0.650 & 0.913 & 0.913 & 0 \\ 
 \hline
 Solar-Energy & 0.010 $\pm$ 0.007 & 1.083 $\pm$ 5.48 & 0.976 & 0.976 & 0 \\ 
 \hline
 NASDAQ & 1.558 $\pm$ 0.460 & 8.953 $\pm$ 8.554 & 0.120 & 0.112 & 7 \\ 
 \hline
\end{tabular}\label{table: HN_128}
\begin{minipage}{13cm}
\end{minipage}
\end{table}

\begin{table}[h!]
\caption{MSE, MAPE, and distance correlation results for all the experiments with a learning rate of 0.001}
\begin{tabular}{||p{0.16\textwidth} | p{0.10\textwidth} | p{0.12\textwidth} | p{0.14\textwidth} | p{0.14\textwidth} | p{0.14\textwidth}||} 
 \hline
 Time Series & MSE & MAPE & $\hat{R}(\textbf{A}_{t}^{(P)}, \textbf{Y})$ (max) & $\hat{R}(\textbf{A}_{T}^{(P)}, \textbf{Y})$ & Change (in \%) \\ [0.5ex] 
 \hline\hline
 AR(1) & 0.028 $\pm$ 0.007 & 0.048 $\pm$ 0.012 & 0.964 & 0.9964 &  0 \\ 
 \hline
 AR(5) & 0.070 $\pm$ 0.036 & 0.099 $\pm$ 0.083 & 0.917 &  0.416 & 55 \\ 
 \hline
 AR(10) & 0.438 $\pm$ 0.184 & 0.272 $\pm$ 0.282 & 0.952 & 0.426 & 55  \\ 
 \hline
 AR(20) & 1.111 $\pm$ 0.250 & 1.259 $\pm$ 2.677 & 0.965 & 0.346 & 64  \\ 
 \hline
 MA(1) & 0.826 $\pm$ 0.060 & 0.367 $\pm$ 0.199 & 0.472 & 0.472 & 0  \\ 
 \hline
 MA(5) & 1.011 $\pm$ 0.078 & 0.399 $\pm$ 0.144 & 0.450 & 0.170 & 62  \\ 
 \hline
 MA(10) & 1.213 $\pm$ 0.092 & 0.468 $\pm$ 0.179 & 0.439 & 0.115 & 74  \\ 
 \hline
 MA(20) & 1.534 $\pm$ 0.130 & 0.569 $\pm$ 0.222 & 0.431 & 0.106 & 75  \\
 \hline
 ARMA(1,1) & 0.008 $\pm$ 0.002 & 0.030 $\pm$ 0.018 & 0.967 & 0.967 & 0  \\ 
 \hline
 ARMA(1,10) & 0.012 $\pm$ 0.004 & 0.037 $\pm$ 0.021 & 0.972 & 0.972 & 0 \\ 
 \hline
 ARMA(10,1) & 0.193 $\pm$ 0.097 & 0.267 $\pm$ 0.359 & 0.945 & 0.508 & 46  \\ 
 \hline
 GARCH(2,2) & 1.201 $\pm$ 1.034 & 1.453 $\pm$ 4.901 & 0.279 & 0.270 & 27  \\ 
 \hline
 GARCH(4,4) & 1.235 $\pm$ 1.201 & 0.581 $\pm$ 0.471 & 0.209 & 0.068 & 20  \\ 
 \hline
    ETTh1, OT & 0.039 $\pm$ 0.021 & 0.523 $\pm$ 0.537 & 0.942 & 0.942 & 0 \\ 
 \hline
 Solar-Energy & 0.010 $\pm$ 0.006 & 1.083 $\pm$ 2.29 & 0.977 & 0.977 & 0 \\ 
 \hline
 NASDAQ & 1.592 $\pm$ 0.497 & 1.132 $\pm$ 0.907 & 0.112 & 0.103 & 8 \\ 
 \hline
\end{tabular}\label{table: LR_001}
\begin{minipage}{13cm}
\end{minipage}
\end{table}

\begin{table}[h!]
\centering
\caption{MSE, MAPE, and distance correlation results for all the experiments with a dropout rate of 0.2}
\resizebox{\textwidth}{!}{
\begin{tabular}{||p{0.16\textwidth} | p{0.13\textwidth} | p{0.13\textwidth} | p{0.14\textwidth} | p{0.14\textwidth} | p{0.10\textwidth}||} 
 \hline
 Time Series & MSE & MAPE & $\hat{R}(\textbf{A}_{t}^{(P)}, \textbf{Y})$ (max) & $\hat{R}(\textbf{A}_{T}^{(P)}, \textbf{Y})$ & Change (in \%) \\ [0.5ex] 
 \hline\hline
 AR(1) & 0.030 $\pm$ 0.009 & 0.060 $\pm$ 0.030 & 0.923 & 0.923 &  0 \\ 
 \hline
 AR(5) & 0.174 $\pm$ 0.160 & 0.184 $\pm$ 0.176 & 0.911 &  0.457 & 50 \\ 
 \hline
 AR(10) & 1.210 $\pm$ 0.470 & 0.610 $\pm$ 0.864 & 0.948 & 0.423 & 55 \\ 
 \hline
 AR(20) & 1.292 $\pm$ 0.291 & 0.774 $\pm$ 1.080 & 0.967 & 0.368 & 62 \\ 
 \hline
 MA(1) & 0.605 $\pm$ 0.037 & 0.378 $\pm$ 0.190 & 0.430 & 0.430 & 0 \\ 
 \hline
 MA(5) & 0.811 $\pm$ 0.047 & 0.437 $\pm$ 0.381 & 0.436 & 0.127 & 71 \\ 
 \hline
 MA(10) & 1.027 $\pm$ 0.056 & 0.492 $\pm$ 0.231 & 0.425 & 0.099 & 77 \\ 
 \hline
 MA(20) & 1.040 $\pm$ 0.056 & 0.550 $\pm$ 0.313 & 0.433 & 0.094 & 78 \\
 \hline
 ARMA(1,1) & 0.015 $\pm$ 0.007 & 0.042 $\pm$ 0.021 & 0.923 & 0.923 & 0 \\ 
 \hline
 ARMA(1,10) & 0.014 $\pm$ 0.005 & 0.044 $\pm$ 0.017 & 0.955 & 0.955 & 0 \\ 
 \hline
 ARMA(10,1) & 0.684 $\pm$ 0.271 & 0.387 $\pm$ 0.363 & 0.948 & 0.499 & 47 \\ 
 \hline
 GARCH(2,2) & 1.128 $\pm$ 0.835 & 0.701 $\pm$ 1.248 & 0.280 & 0.276 & 2 \\ 
 \hline
 GARCH(4,4) & 1.226 $\pm$ 1.252 & 1.666 $\pm$ 4.024 & 0.226 & 0.213 & 6 \\ 
 \hline
    ETTh1, OT & 0.041 $\pm$ 0.019 & 0.842 $\pm$ 2.136 & 0.940 & 0.940 & 0 \\ 
 \hline
 Solar-Energy & 0.029 $\pm$ 0.031 & 0.663 $\pm$ 1.47 & 0.985 & 0.985 & 0 \\ 
 \hline
 NASDAQ & 1.562 $\pm$ 0.487 & 2.514 $\pm$ 4.312 & 0.096 & 0.088 & 9 \\ 
 \hline
\end{tabular}\label{table: DO_02}
}
\end{table}

\clearpage

\bibliographystyle{elsarticle-num} 
\bibliography{refs}

\end{document}